\documentclass[acmsmall,screen,nonacm]{acmart}
\acmJournal{TELO}
\AtBeginDocument{%
  }

\usepackage{caption}
\usepackage{subcaption}
\usepackage{algorithmic}
\usepackage{algorithm}
\usepackage{multirow}
\usepackage{array}
\begin{document}

\title{Pareto Set Prediction Assisted Bilevel Multi-objective Optimization }

\author{Bing Wang}
\email{bing.wang@unsw.edu.au}
\affiliation{%
  \institution{School of Engineering and Technology, University of New South Wales Canberra}
 \country{Australia}
}

\author{Hemant K. Singh}
\email{h.singh@unsw.edu.au}
\affiliation{%
  \institution{School of Engineering and Technology, University of New South Wales Canberra}
\country{Australia}
}

\author{Tapabrata Ray}
\email{t.ray@unsw.edu.au}
\affiliation{%
  \institution{School of Engineering and Technology, University of New South Wales Canberra}
\country{Australia}
}

\renewcommand{\shortauthors}{}

\begin{abstract}
 Bilevel optimization problems comprise an upper level optimization task that contains a lower level optimization task as a constraint. While there is a significant and growing literature devoted to solving bilevel problems with single objective at both levels using evolutionary computation, there is relatively scarce work done to address problems with multiple objectives~(BLMOP) at both levels. For black-box BLMOPs, the existing evolutionary techniques typically utilize nested search, which in its native form consumes large number of function evaluations. In this work, we propose to reduce this expense by predicting the lower level Pareto set for a candidate upper level solution directly, instead of conducting an optimization from scratch. Such a prediction is significantly challenging for BLMOPs as it involves one-to-many mapping scenario. We resolve this bottleneck by supplementing the dataset using a helper variable and construct a neural network, which can then be trained to map the variables in a meaningful manner. Then, we embed this initialization within a bilevel optimization framework, termed Pareto set prediction assisted evolutionary bilevel multi-objective optimization (PSP-BLEMO). 
Systematic experiments with existing state-of-the-art methods are presented to demonstrate its benefit. The experiments show that the proposed approach is competitive across a range of problems, including both deceptive and non-deceptive problems~\footnote{This is author submitted version of the work currently undergoing peer-review}.
\end{abstract}




\maketitle

\section{Introduction}
A number of real-world problems manifest as a hierarchical optimization problem, where the objective(s) of an upper-level~(UL) problem are optimized subject to the optimality of a lower-level~(LL) problem. The UL and LL are also referred to as leader and follower tasks, respectively, and may contain their own additional constraints. Such problems are referred to as bilevel programming or bilevel optimization problems~(BLP)~\cite{sinha2018review}; and are the most-studied subset of a more generalized class of \emph{multi-level} optimization problems~\cite{lu2016multilevel}. Such problems are of interest in a number of application domains~\cite{sinha2018review}, including transportation\cite{calamai1994generating}, economics~\cite{fudenberg1991game,sinha2014finding}, engineering~\cite{kirjner1998outer} and management~\cite{bard1983coordination}, to name a few. Moreover, they also pose certain unique theoretical and algorithmic challenges~\cite{islam2017surrogate} compared to traditional optimization~(referred to occasionally in this study as \emph{single-level} optimization for specificity). These include, for example, being NP-hard even when the problems are linear at both levels~\cite{sinha2018review}, deceptive nature of UL evaluations for solutions that are sub-optimal at LL, and challenges in environmental selection and benchmarking strategies~\cite{islam2017surrogate,mejia2022pseudo}. Considering the above, bilevel problems have attracted interest from both academic researchers and practitioners. Formally, bilevel problems can be represented as shown in Eq.~\ref{eq:bldef}. 

\begin{equation}\footnotesize
\label{eq:bldef}
\begin{array}{lr}
 \underset{\mathbf{x}_u\in \mathbb{X}_u}{\text{Minimize}}\hspace{2mm} F(\mathbf{x}_u,\mathbf{x}_l) = F^1(\mathbf{x}_u, \mathbf{x}_l), ...., F^M(\mathbf{x}_u, \mathbf{x}_l) \\ 
\text{Subject to:} \\ 
\hspace{3mm} \mathbf{x}_l \in \underset{\mathbf{x}_l\in \mathbb{X}_l}{\text{argmin}} \hspace{2mm} f(\mathbf{x}_u, \mathbf{x}_l) =f^1(\mathbf{x}_u,\mathbf{x}_l) \ldots f^N(\mathbf{x}_u,\mathbf{x}_l)\\ 
\hspace{10mm} \text{subject to: } g^1(\mathbf{x}_u, \mathbf{x}_l),g^2(\mathbf{x}_u, \mathbf{x}_l)\ldots g^q(\mathbf{x}_u, \mathbf{x}_l) \leq 0\\
\hspace{3mm} G^1(\mathbf{x}_u, \mathbf{x}_l), G^2(\mathbf{x}_u, \mathbf{x}_l)\ldots G^p(\mathbf{x}_u, \mathbf{x}_l) \leq 0\\
\end{array}
\end{equation}
\noindent Here $F^1 \ldots F^M$ and $f^1 \ldots f^N$ are objective functions at UL and LL, respectively. The UL and LL design variables are denoted using $\{\mathbf{x}_u, \mathbf{x}_l\}$, sampled from their design spaces $\{\mathbb{X}_u, \mathbb{X}_l\}$, respectively. The function $g(\mathbf{x}_u, \mathbf{x}_l)$ represents a constraint for the LL problem, while $G(\mathbf{x}_u, \mathbf{x}_l)$ determines feasible space for the UL problem. For non-exact techniques, the equality constraints, if they exist, are often converted into inequalities with a tolerance, hence equality constraints are not included in Eq.~\ref{eq:bldef} for brevity. 

Bilevel optimization problems have their origins in Stackelberg game theory and economics~\cite{sinha2018review}, but have been since studied in several other fields. However, overwhelming majority of these studies have been dedicated to problems where both levels have a single objective~(BLSOP), i.e., $M=N=1$. A number of competitive algorithms have been developed for this class of problems utilizing ideas such as problem transformation~\cite{sinha2018review}, hybridization~\cite{islam2017enhanced}, surrogate-assisted search~\cite{islam2017surrogate,sinha2021solving}, transfer learning~\cite{bing2022cec,LeiChen2021}. For more comprehensive overview of the works in the area of BLSOP, the readers may refer to review papers such as \cite{sinha2018review,lu2016multilevel,talbi2013taxonomy}. 

However, there has been relatively scarce attention paid to bilevel problems with multiple objectives at one or both levels. The bilevel multi-objective optimization problems~(BLMOPs), also denoted with acronym MOBO in \cite{mejia2023multiobjective}, typically have multiple objectives at UL, while the LL may have one or multiple objectives. The special case where the UL has a single objective, but LL has multiple objectives, is referred to as semi-vectorial bilevel optimization~(SVBO) problem~\cite{alves2016illustration,Halter2006,andreani2019bilevel}. A number of real-world problems can be formulated as BLMOPs, and thus could benefit from efficient methods to solve such problems. In \cite{sinha2015towards,islam2017enhanced}, an environmental economics problem is discussed wherein a company wants to set up a gold mine in a specific geographical region. The government acts as the UL decision-maker in this case, with the objectives of maximizing revenues generated by the project~(tax, jobs etc.) and minimizing the detrimental impacts on the environment. The mining company, as the LL decision-maker, aims to maximize its profits as well as maximize its reputation/public image. Another problem discussed in \citep{sinha2015towards,islam2017enhanced} is that of hierarchical decision-making in a company, wherein the objectives of a chief executive officer~(CEO) are to maximize the quality of the products and the company profits. On the other hand, the branch heads working at a lower level seek to maximize the branch profits and the worker satisfaction. In the context of border security, the problem discussed in \cite{Lessin2019sensor} involves UL objectives of maximizing total weighted exposure of the minimal exposure path, minimizing sensor relocation time and the number of sensors relocated, while the LL seeks to minimize the total expected weighted exposure of the intruder’s minimal exposure path. Some other BLMOP application examples include  transportation planning and management~\cite{yin2002multiobjective}, manufacturing~\cite{gupta2016pareto}, machine learning problems~\cite{kamani2020targeted, ozdayi2021fair}. A number of additional applications from the domains of logistics, environmental economics and manufacturing can also be found in \cite{mejia2023multiobjective}.

In this study, we are predominantly concerned with addressing BLMOPs where both levels have multiple objectives. Given the limited work in the field so far, the scope of study is limited to two objectives, but in principle higher number of objectives fall within the same category. Beyond the challenges already encountered for BLSOPs, BLMOPs have their own characteristic challenges. The key challenge is that for any given UL solution, the optimum solution at LL is not just a single solution but a set of trade-off solutions~(Pareto front, PF). This situation occurs rarely in BLSOPs, where the LL optimum has multiple global optimum solutions, but is ubiquitous for BLMOPs. This implies that for each $\mathbf{x}_u$, not only a significant number of function evaluations are required at LL to find a good PF approximation, a large set of solutions~(corresponding to LL PF) also needs to be evaluated at UL. This results in proliferation of function evaluations used by approaches, especially those that rely on nested strategies. The second, related challenge is that of autonomy in decision-making by UL and LL. Since LL decision-maker has an entire PF approximation to chose from for sending to UL decision-maker, co-operation or conflict between the two decision-makers may result in sub-optimal PF at the UL. In most of the existing works, an optimistic version of the problem is assumed, wherein the UL decision-maker has the authority to choose solutions from the LL PF that bring UL the most benefits. In what follows, we highlight some of the representative approaches that address BLMOPs. Moreover, some of the works convert the LL problem from multi-objective to single-objective through the use of value functions in order to provide a single solution to the UL~\cite{sinha2015towards}. For a more extensive coverage of literature in BLMOPs, the readers are referred to ~\cite{mejia2023multiobjective,sinha2013bilevel}.

\textbf{\textit{Classical/analytical techniques: }} 
Much like the case of BLSOPs, some of the early efforts were directed towards solving BLMOPs utilizing mathematical techniques for exact solutions. The pre-condition to the application of these techniques is often that the underlying response functions~(objectives, constraints) need to satisfy certain regularity conditions. For instance, in \cite{nishizaki1999stackelberg}, an approach is developed to solve BLMOPs where all objectives and constraints at both levels are linear. Three different cases were solved with regards to anticipated behavior of the LL decision-maker, namely, optimistic, pessimistic, and historical. In \cite{OSMAN2004239,abo2001bi}, bi/multi-level problems with multiple objectives are solved under a scenario where the decision-makers have certain tolerances, modeled using fuzzy set theory. In \cite{shi2001model, shi1997interactive}, an interactive method was presented for BLMOP by replacing the LL problems with Kuhn–Tucker conditions. In \cite{eichfelder2010multiobjective}, an approach to solve non-linear BLMOPs was presented. Differentiability is generally assumed for calculating gradients in such approaches.
In~\cite{pieume2011solving}, BLMOP is linearly scalarized to generate Pareto frontier solutions, and then a filtering mechanism is constructed to maintain distribution for these solutions. Linear functions are assumed for the given problems.  In~\cite{alves2012computing}, Pareto solutions of bilevel linear problems have been characterized after converting bilevel problem into single level mixed 0-1 programming problems. 
Moreover, in most of these works, the working principles are demonstrated numerically on a very limited set of problems~(often one or two), with small number of variables. 

\textbf{\textit{Metaheuristic techniques: }} In order to handle cases where the underlying functions do not satisfy the required mathematical properties, or are not available altogether~(so-called ``black-box'' functions), metaheuristic techniques, such as evolutionary algorithms or swarm intelligence algorithms, are often chosen. However, this versatility typically comes with the requirement to evaluate large number of candidate solutions to reach near-optimal solutions. Given the structure of a bilevel problem, a common way to solve it using metaheuristic methods is through a nested approach, where a standard multi-objective evolutionary algorithm~(MOEA) or equivalent methods are used at both levels. Towards this end, non-dominated sorting algorithm II~(NSGA-II) with a special population structure was utilized in \cite{deb2009solving} in a nested manner to solve BLMOPs. This was further hybridized with local search methods in \cite{deb2010efficient}. Differential evolution was implemented in a nested form for solving BLMOPs in \cite{islam2016nested}.
With similar intent, nested search has also been proposed using other metaheuristic approaches, such as particle swarm optimization~(PSO) and simulated annealing~ (SA)~\cite{zhang2012improved, zhang2016improved}.
Recently, a variable decomposition based cooperative co-evolutionary method was proposed to improve search efficiency of nested search structure of BLMOP~\cite{cai2022cooperative}. 
A generic framework to solve BLMOPs was also presented recently in \cite{mejia2022novel}, wherein some gain in efficiency was realized via a representation that utilizes grouping of certain variables as \emph{families}. 

Since the implementation of meta-heuristic search in a nested mode involves, in particular, a large number of LL function evaluations~(FE), a pertinent effort is to develop strategies to expedite the search at LL and reduce LL FE. Towards this end, transfer of solutions obtained corresponding to a neighboring UL solution as a seeding population has been investigated for BLSOPs in \cite{ankur2017bleaq},\cite{LeiChen2021},\cite{bing2022cec}, and more recently, extended for BLMOPs in~\cite{wang2023blmotrans}. Though the results were promising, the performance of such simple transfer strategies is subject to neighboring landscapes being similar, as was demonstrated for BLSOPs in \cite{bing2022cec}. Another promising direction to improve LL search efficiency is by generating LL optimal solutions through certain prediction mechanism. Often, the LL optimal solutions are not entirely independent of each other, since they are related by the UL variables that act as parameters for the LL problems. Consequently, by learning from the dataset of previously evaluated solutions, predictions can be made regarding the optimal $\mathbf{x}_l$ for a new candidate $\mathbf{x}_u$. The predicted LL optimum can be used as a starting point for local search, or to seed initial population to reduce the effort required in carrying out the LL optimization. In mf-BLEAQ~\cite{sinha2017approximated}, a quadratic fiber based surrogate model was proposed to predict LL Pareto sets (PS). Using predicted solutions to start the LL search provided a head-start. However, this modeling scheme needs to build a model for each variable of a solution, and for LL PS there is a need for multiple solutions to cover PF. Therefore, a large number of models are required to be built and maintained. A more recent study~\cite{Conditional_wang2023} proposes to reduce the number of prediction models for LL PS to a single model, i.e., conditional generative neural network~(cGAN). Relying on the approximation capacity of cGAN, the proposed algorithm cG-BLEMO only needs to maintain one model. However, due to Gaussian noise introduced in cGAN, LL solutions generated by cGAN do not follow the distribution of the PS/PF closely, requiring additional search to improve the solutions to capture LL PS. Furthermore, to keep the training time practicable for cGAN, the datasize used had to be kept relatively small in the implementation~(typically up to 800).

Continuing the above line of inquiry, in this study, we aim to advance the research direction that uses LL PS prediction for solving BLMOPs efficiently. Towards this goal, we build a simple yet accurate model customized for BLMOPs, that can be used to predict LL PS in lieu of evolutionary search to drive majority of the search. The key contributions of this work can be summarized as:
\begin{itemize}
    \item Ordinarily, the LL optimum comprises multiple solutions~(PS) for each UL solution $\mathbf{x}_u$. Thus, the dataset of previously evaluated solutions results in a one-to-many mapping with regards to LL PS, which is unsuitable for building prediction models. We propose a simple transformation of the data to make them amenable for creating one-to-one mapping. 
    \item We then use simple feedforward neural networks to build the prediction model that maps $\mathbf{x}_u$ and optimal $\mathbf{x}_l$~(LL PS). The model training is relatively fast, uses fewer hyperparameters compared to more sophisticated models previously used in the literature, and is more accurate in terms of its LL PS prediction. 
    \item We embed the above PS prediction into the nested search for BLMOPs. After accumulation of training data and construction of the NN model, LL search can gradually be delegated to prediction. To maintain/improve the prediction accuracy, NN is re-trained periodically during the run with new evaluated solutions.
    \item We benchmark the proposed approaches on a diverse set of BLMOPs from the literature, including \emph{deceptive} problems that have been scarcely studied. Comparison with state-of-art algorithms highlights its competent performance in terms of proximity to optimum, and efficiency in terms of function evaluations used.   
\end{itemize}

The remainder of this paper is structured as follows. In Section~\ref{sec:background}, we present the main proposed idea for PS prediction and show proof-of-concept results to establish its viability. 
Then, this prediction method is embedded within a nested bilevel multi-objective optimization framework, the details of which are presented in Section~\ref{sec:proposed}. The numerical experiments and discussions are presented in Section~\ref{sec:experiment}, followed by concluding remarks in Section~\ref{sec:conclusion}.

\section{The Main Idea and Proof-of-concept}
\label{sec:background}

\begin{figure*}[!ht]
    \centering
     \includegraphics[width=0.9\textwidth]{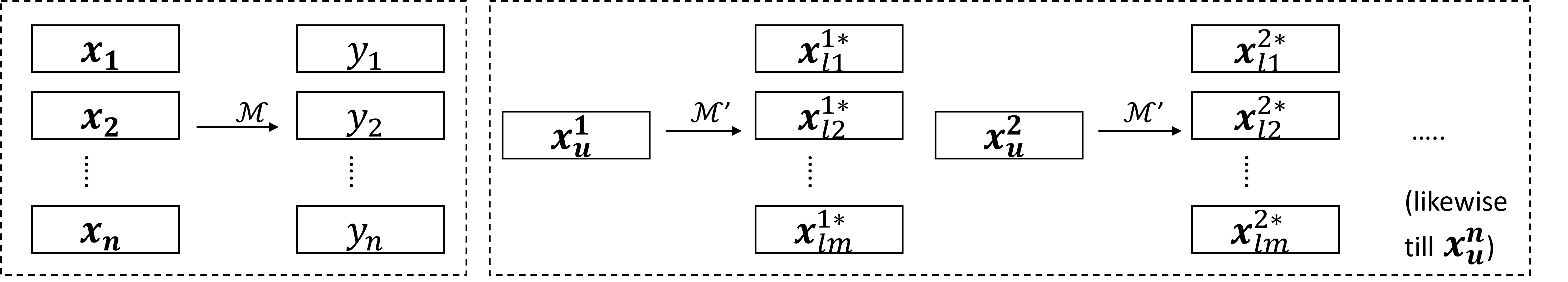}
    \caption{The difference between standard mapping~(left), and the bilevel mapping of interest in this study~(right). On the left, commonly used surrogate modeling has one-to-one/many-to-one mapping between inputs and outputs, where a vector ${\mathbf{x}}$ corresponds to one value of $y$. On the right, the mapping involves  a set of vectors~(the LL PS $\mathbf{x}_{l}^{i*}$) corresponding to each given candidate UL vector $\mathbf{x}_u^i$, resulting in a one-to-many mapping scenario. }
    \label{fig:illus_M}
\end{figure*}

In standard (single-level) multi-objective optimization, it is common to seek intermittent estimation of Pareto front/Pareto sets to aid the search, especially when the underlying response functions~(objectives, constraints) are computationally expensive in nature. This can be done by training a surrogate model on an archive of evaluated solutions to establish a mapping between the input $\mathbf{x}_i$ and output response(s) $y_i$, as shown on the left side in Fig.~\ref{fig:illus_M}. Thereafter, an internal optimization can be conducted using the surrogate model to come up with solutions that are likely to be Pareto optimal, which can then undergo the expensive evaluation~\cite{shankar2016multi}. Given that the accuracy of the models is dependent on the dataset, this process usually needs to be conducted multiple times over a run. More recently, there have also been attempts~\cite{lin2022pareto,lin2020controllable,deb2023learning,ma2020efficient,navon2021learning} to predict the Pareto optimal front based on preference vectors. This is done by utilizing preferences~(quantified using, e.g., reference vector coefficients) with regards to different objectives as inputs, and certain known Pareto optimal solutions as output. The mapping between these can be done using various machine learning models such as neural networks, Gaussian process or hypernetworks. The mapping can then be used to generate new solutions that are likely to be Pareto optimal; which can be used, for example to increase the density of solutions on the PF, or aid the search interactively.

For the above modeling tasks, the dataset usually has a one-to-one or many-to-one relationship between inputs and outputs, for which there is no mathematical ambiguity in terms of building the mappings. In BLMOPs however, this is mostly not the case. If the models are built individually for LL or UL functions to aid the search independently, the above models and types of mappings can still be maintained. However, to potentially bypass the LL optimization for some UL candidate solutions entirely by predicting the LL PS, a different type of mapping is required. As shown in the right side of Fig.~\ref{fig:illus_M}, the available dataset in this case may be the previously evaluated solutions, $\mathbf{x}_{u}^i$. For \emph{each} of these solutions, a lower level PS approximation has been found using an MOEA, denoted as $\{\mathbf{x}_{lj}^{i*}$,$j=1\ldots m \}$. That task is to predict the PS $\{\mathbf{x}_{l}^* \}$ directly for a new $\mathbf{x}_{u}$ sampled during the UL search. For BLSOPs, in most cases this can still be  modeled as a one-to-one mapping~\cite{sinha2018review}, but for BLMOPs, it is evident that such a mapping is not straightforward due to the same value of input $\mathbf{x_{u}}$ for multiple values of $\mathbf{x}^*_{l}$.

To address some of the issues above, we propose a simple method in this work to transform the data and use neural networks~(NN) for creating a mapping between $\mathbf{x}_{u}$ and $\mathbf{x}^*_{l}$. Neural networks can inherently support vectors as outputs~(which is the case with $\mathbf{x}^*_{l}$), and are also proven to be universal approximators given sufficient training data. Given the number of points accumulated with LL PS corresponding to multiple $\mathbf{x}_{u}$ values, the size of the dataset works in favor of neural networks for BLMOPs to create accurate mappings. Of course, the issue of one-to-many mapping still needs to be resolved. We use a helper variable to supplement the input data to overcome this. Our proposed approach is quite simple in its structure and discussed below with an example. 

\begin{figure*}[!ht]
    \centering
     \includegraphics[width=0.9\textwidth]{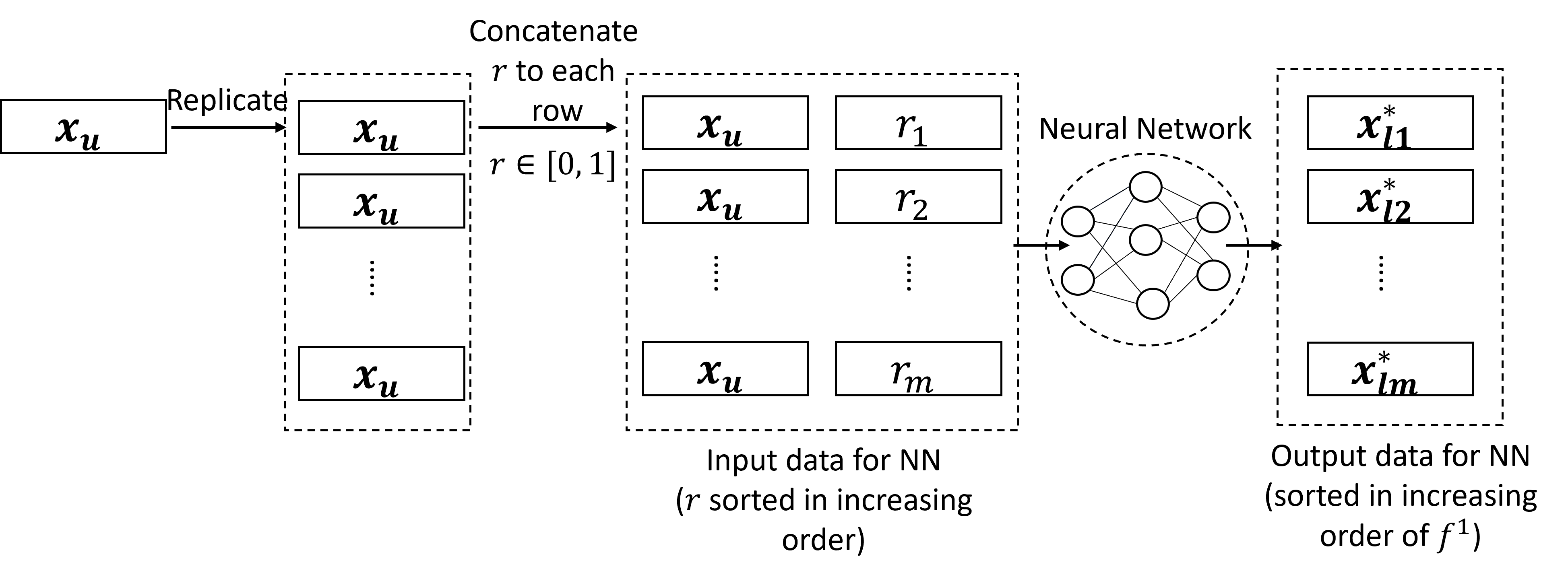}
    \caption{The data for all available $\mathbf{x}_{u}$ and their corresponding LL PS is processed in this manner and then stacked together as one collective dataset provided to the NN for training.}
    \label{fig:illus_data}
\end{figure*}

To begin with, the dataset that will be provided to the NN needs to be generated and conditioned. To generate the initial data, the LL PS needs to be identified for certain initial samples of $\mathbf{x}_{u}$. These LL PS (or approximations thereof) can be obtained using an MOEA at the LL. Once obtained, the available data resembles the one shown on the right in Fig.~\ref{fig:illus_M}. Then, the data for each $\mathbf{x}_{u}$ is conditioned schematically as shown in Fig.~\ref{fig:illus_data}. To do so, the LL PS approximations $\{\mathbf{x}_l^*\}$ are first sorted based on increasing order of one of the LL objectives; chosen (without loss of generalization) as $f^1$ in this case. This ordered set forms the output values for the NN. For the inputs, the value of a given $\mathbf{x}_{u}$ is replicated $m$ times, where $m$ is the number of solutions in the LL PS approximation corresponding to $\mathbf{x}_{u}$. In addition, a helper variable $r$ is appended to the inputs. The variable $r$ assumes a uniformly spaced values in the range of 0 to 1, where $r_1$ corresponds to 0 and $r_m$ corresponds to 1. The $r$ values can thus be generated easily as per the number of points in the LL PS approximation. As it can be seen, the resulting dataset has unique input vectors, each mapping to a single output vector $\mathbf{x}^*_{l}$. Therefore, the dataset is suitable for many-to-one or one-to-one mapping. The significance of having $\mathbf{x}^*_{l}$ in the sorted order of $f^1$, which corresponds to sorted order of $r$ in the inputs, can be inferred as providing a sense of direction. This can be thought of as akin to mapping solutions along different preferences or reference vectors in, e.g., decomposition-based evolutionary algorithms. It is worth noting that, using a single helper variable $r$ is only suitable for solving bi-objective problems, because there is an unambiguous spatial ordering among the PF points for bi-objective problems. That is, the PF points can be traversed from one corner of the PF to another, corresponding to the ordered set of reference vectors based on the monotonic progression of $r$ values. Such ordering is not straightforward for problems with more than two objectives. Using the proposed approach for more than two objectives may be feasible following similar ideas of decomposition, but will require (a) more than one helper variables, and (b) careful consideration of the ordering of the data points and helper variables. Given that most of the existing BLMOPs involve two objectives~\cite{deb2010efficient} and that the existing methods require significant developments even for bi-objective problems, we limit the scope of this study to two objectives.

Once the above processing has been applied to all available $\mathbf{x}_{u}$ and the corresponding $\mathbf{x}^*_{l}$, the data is stacked to form the dataset for training a feed-forward NN~(FNN). It takes the inputs~($\mathbf{x}_{u},r$) and builds a model $\mathbb{M}'$ to predict the output $\mathbf{x}^*_{l}$. This model can then be integrated in the bilevel search method to predict the LL PS for any new candidate $\mathbf{x}_{u}$ under consideration, or can form a seed population to expedite the LL search towards its PS. 

The benchmark problem DS2~\cite{deb2010efficient}, with two variables at each level, is used here for proof-of-concept. The theoretical LL PS can be inferred for the problem for any given $\mathbf{x}_u$, which can be used to verify the predication accuracy. For this illustration, we sample $10$ random solutions~($\mathbf{x}_u$) at the UL. For each $\mathbf{x}_u$, we sample $20$ Pareto solutions~($\mathbf{x}^*_l$) on the LL PS. This data is then supplemented with the helper variable values as discussed above, and the FNN is trained. 

\begin{figure}[!ht]
    \centering
    \subfloat[Predicted PF with ordered $r$]  
    {\includegraphics[width=0.3\textwidth]{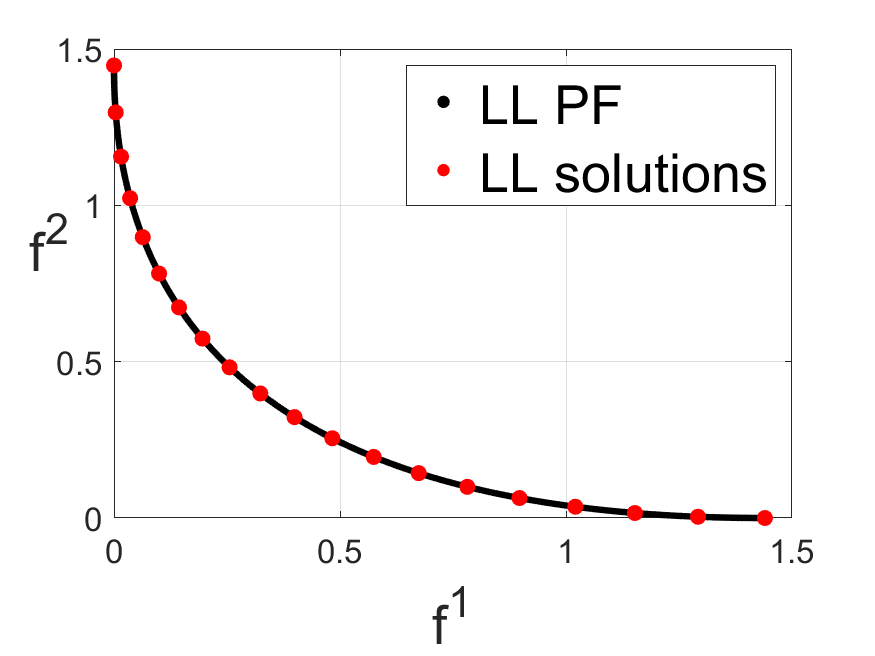}
         \label{fig:ps_withr}} 
     \subfloat[Predicted PF with random $r$]
    {\includegraphics[width=0.3\textwidth]{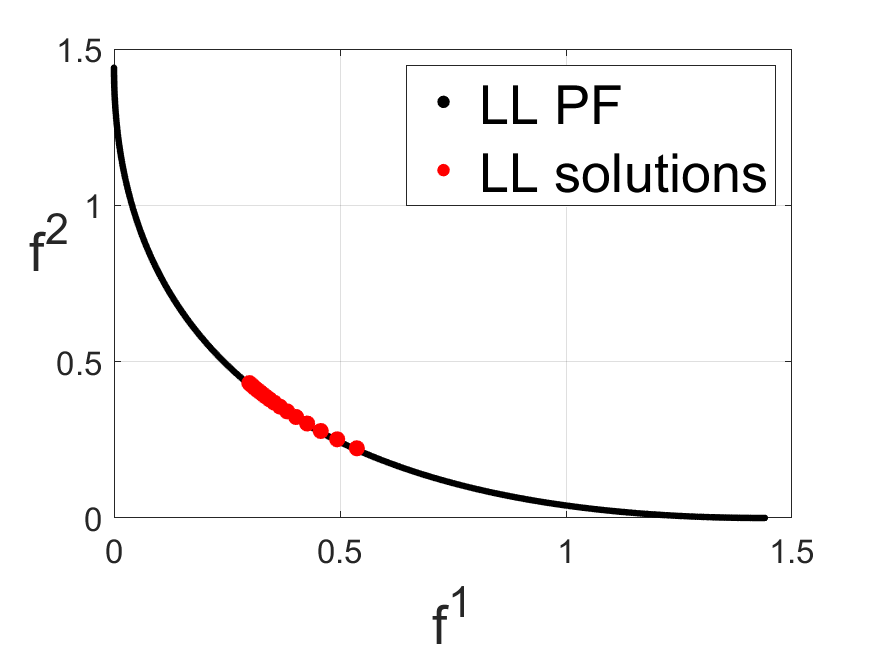}
        \label{fig:ps_withoutr}}
    \subfloat[Random LL initialization]
    { \includegraphics[width=0.3\textwidth]{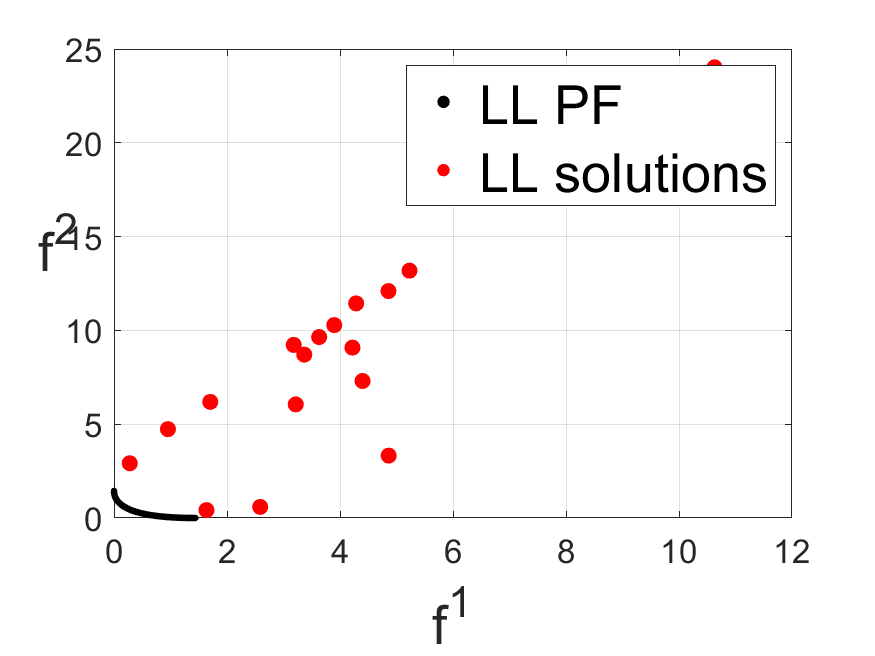}
      \label{fig:normal_sampling}} \\
    \subfloat[Predicted PS with ordered $r$]{
         \includegraphics[width=0.3\textwidth]{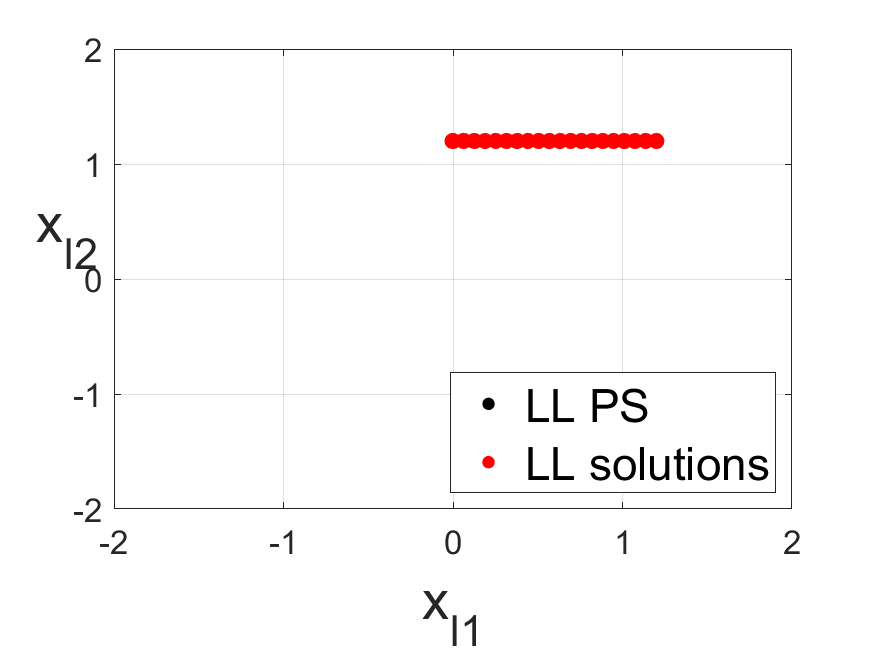}
         \label{fig:ps_withr_x}}    
    \subfloat[Predicted PS with random $r$]{
     \includegraphics[width=0.3\textwidth]{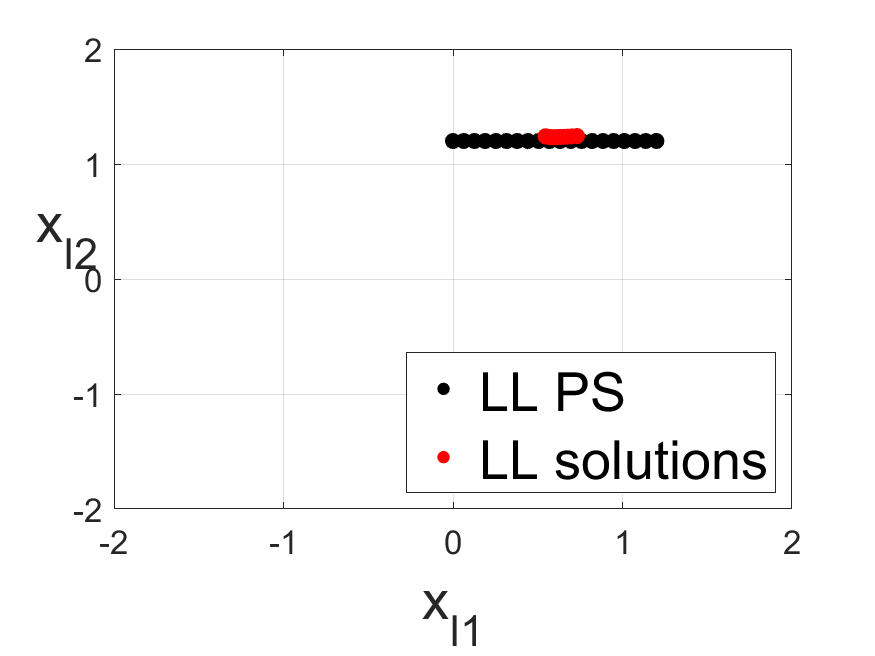}
     \label{fig:ps_withoutr_x}}    
    \subfloat[Random LL initialization]{
         \includegraphics[width=0.3\textwidth]{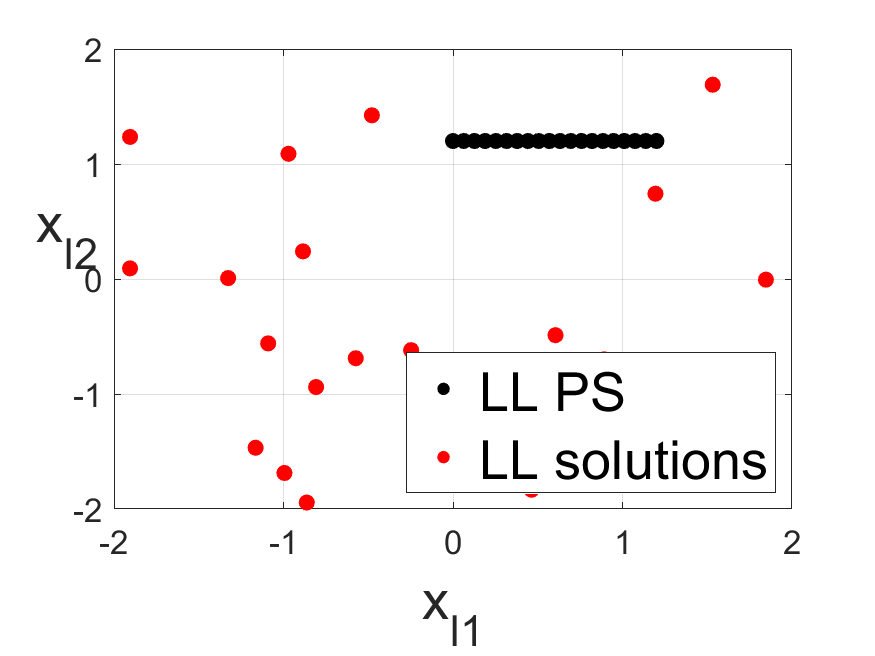}
         \label{fig:normal_sampling_x}}
    \caption{Proof-of-concept results for DS2 problem}
    \label{fig:psgenerator_compare}
\end{figure}

We use MATLAB feedforward neural network tool (\texttt{feedfowardnet}) to create the mapping from $\{(\mathbf{x}_u, r)\}$ to $\{\mathbf{x}^*_l\}$. One hidden layer with 4 nodes is set for this problem. Default parameter settings are used for the training step, mean square error is used as loss function, and backpropagation is used for training (More detailed parameters will be discussed in Section IV). Next, we generate a new $\mathbf{x}_u$ and attempt to predict its PS. The generated UL variable, $\mathbf{x}_u=[1.2, 1.2]$ is combined with  $r$ values  to predict 20 LL PS solutions. The image of predicted PS in the LL objective space is shown in Fig.~\ref{fig:ps_withr}. 

It can be seen that predicted solutions lie on the true PF as well as exhibit a good diversity on it. In Fig.~\ref{fig:ps_withr_x}, in variable space, we plot PS and predicted solutions. It shows consistency between PS and predicted solutions, For comparison, we also show the results obtained when the $r$ is not appended to $\mathbf{x}_u$ in a uniformly increasing sequence but randomly generated instead~(this  is also equivalent to not sorting the $\mathbf{x}^*_l$ based on $f^1$). The results can be seen in  Fig.~\ref{fig:ps_withoutr} and Fig.~\ref{fig:ps_withoutr_x}, where the mapping is able to predict solutions close to only a small part of the PF and PS, respectively. Lastly, we present $20$ solutions that are randomly sampled in the LL search space for the same $\mathbf{x}_u$. This is shown in Fig.~\ref{fig:normal_sampling}, and highlights that if the LL search was conducted in a standard manner starting from a random population, the initial solutions will be far away from the PF. Instead, if the predicted PS such as that in Fig.~\ref{fig:ps_withr} is utilized, the search can be bypassed entirely for some generations, or seeded on/close to the PS, expediting the LL search and potentially saving significant number of LL evaluations. Using this central idea, we build an algorithmic framework to solve BLMOPs, termed Pareto set prediction assisted bilevel evolutionary multiple objective optimization algorithm~(PSP-BLEMO) next.

\section{Proposed algorithm}
\label{sec:proposed}

\begin{figure}
    \centering
    \includegraphics[width=0.7\textwidth]{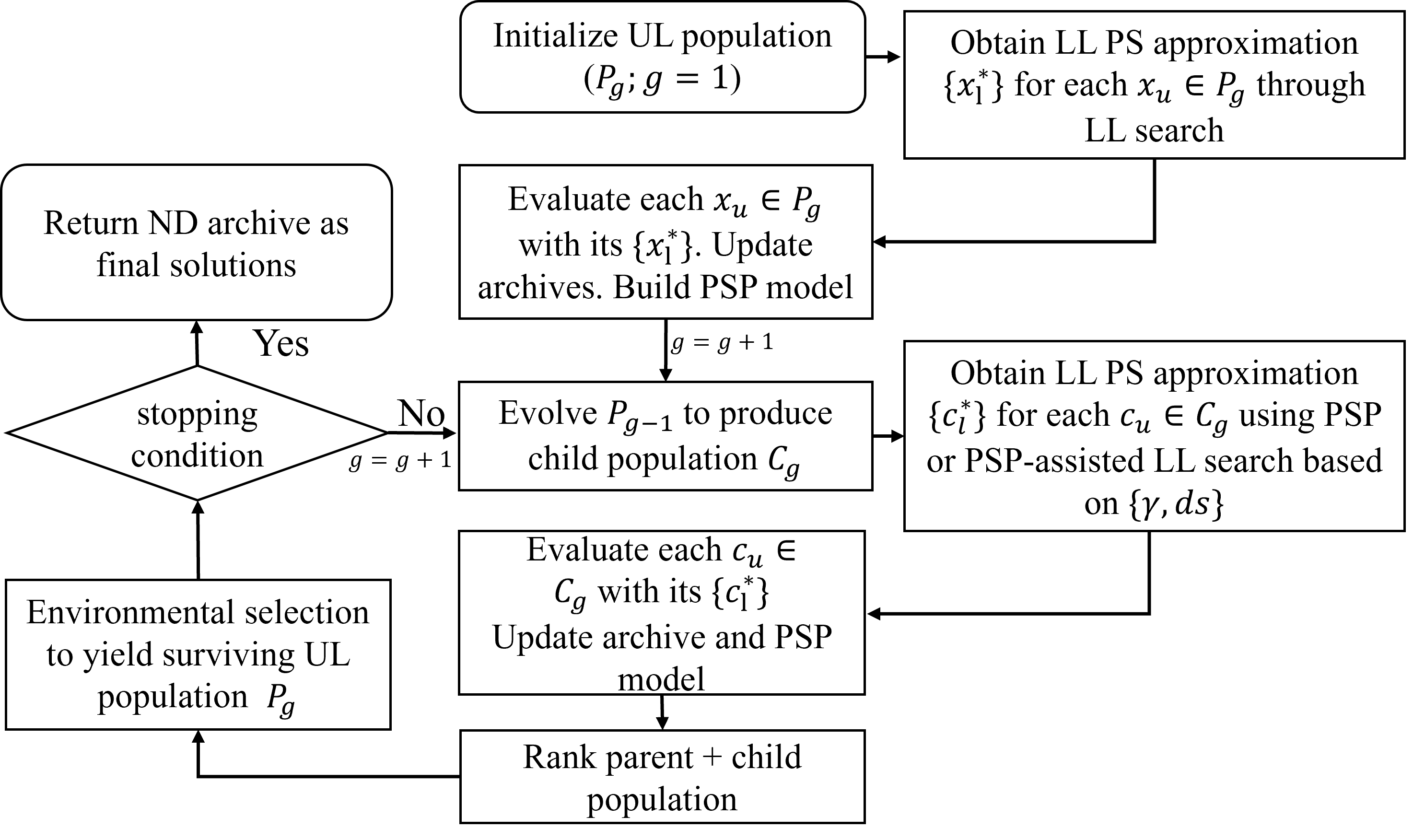}
    \caption{Flowchart of the proposed PSP-BLEMO framework}
    \label{fig:flowchart}
\end{figure}

The general framework of PSP-BLEMO\footnote{The code and data will be made available for research purpose after the review process.} is outlined in Algo.~\ref{algo:blmo_framework} and Fig.~\ref{fig:flowchart}, followed by description of its key components in the following subsections. At UL, the search is conducted through an MOEA, while the LL search has the PSP model integrated in it. 

\begin{algorithm}[!ht] \footnotesize
\caption{PSP-BLEMO framework} 
\begin{flushleft} 
\textbf{Input:} Population size $n^u$;  Evolutionary parameters, NN update parameter $\gamma$, NN data size limit $ds$ \\
 \textbf{Output:} PF/PS approximation for the UL problem
\end{flushleft}
\begin{algorithmic}[1]
\STATE Check LL variable association, represented using binary vector $v$ \label{line:vaa_process}
\STATE Initialize UL population $P_1 = \{\mathbf{x}_u^1,...,\mathbf{x}_u^i,... \mathbf{x}_u^{n^u}\}$
\STATE Identify LL solutions $\{\mathbf{x}_l^{i*}\}$ for  each $\mathbf{x}_u^i$ using Algo.~\ref{algo:causal_LL_search} \label{line:LLsearch_causalcheck}
\STATE Evaluate $\mathbf{x}_u^i$ with feasible $\{\mathbf{x}_l^{i*}\}$ to obtain population fitness $P_1^F$.
\STATE Initialize the ND archive using the $P_1$ \label{framework:archive1}
\STATE Update training data and PSP model; Set generation counter $g=2$ \label{framework:nn_update}
\WHILE{termination condition not met}
\STATE Apply evolutionary operators to $P_{g-1}$ to generate child population $C_g = \{\mathbf{c}_u^1, \ldots \mathbf{c}_u^i, \ldots\mathbf{c}_u^{n^u} \}$. Discard previously evaluated solutions and duplicate solutions from $C_g$\label{framework:child_generation}
\IF{$mod(g,\gamma)==0$ or $ds$ not met} \label{line:gamma}
\STATE Apply PSP assisted search (Algo.~\ref{algo:generator_assisted_LLsearch}) on LL to identify $\{\mathbf{c}_l^{i*}\}$ for each $\mathbf{c}_u^{i}$
\STATE Update PSP model with most recent (up to) $ds$ data points\label{framework:selective_update}
\ELSE
\STATE Use PSP model to predict $\{\mathbf{c}_l^{i*}\}$ for each $\mathbf{c}_u^{i}$ \label{framework:LL_skip}
\ENDIF
\STATE Evaluate $\mathbf{c}_u^i$ with $\{\mathbf{c}_l^{i*}\}$ to obtain fitness $C_g^F$; Update ND archive \label{framework:eval_child}
\STATE Rank the combined population $P_{g-1}^F \cup C_g^F$  \label{framework:sorting}
\STATE Select the surviving population $P_g$ through environmental selection~(Algo.~\ref{algo:environment_selection}); Increment generation count $g = g + 1$ \label{line:env-select1}
\ENDWHILE
\STATE Return ND archive as the solution to UL problem
\end{algorithmic}
\label{algo:blmo_framework}
\end{algorithm}

\subsection{UL search and variable association check}
The UL search commences with an initial LL variable association check, followed by the creation of an initial population $P_1$
generated through a random uniform distribution. An LL variable association check is done for accommodating an uncommon feature of some BLMOPs, referred to as variable association ambiguity~(VAA)~\cite{bing2024BL}. If a BLMOP has LL VAA, a subset of its LL variables participate solely in UL objectives. A detailed discussion of VAA can be found in \cite{bing2024BL}, here we only briefly introduce the VAA checking step adapted to this study, to keep the primary focus of this study on PSP module. The VAA check is done by perturbing an LL variable while holding all other variables constant. By comparing the LL objective values before and after the perturbation, it becomes possible to ascertain whether this variable is associated with LL objectives. Likewise, the UL objective values are also checked before and after the perturbation to ascertain which LL variables affect them. This process is repeated for all LL variables one by one. Subsequently, a binary vector $v$ (Algo~\ref{algo:blmo_framework}, Line~\ref{line:vaa_process}) is generated to record the association status of each variable: 0 signifies no VAA, while 1 indicates otherwise.\footnote{Note that theoretically another case is possible where an LL variable is completely redundant, i.e, does not feature in LL or UL. We have omitted discussion of such cases for brevity in this study, but they can be easily filtered out from the search entirely through the above VAA checks.}
Upon completion of the VAA check and the generation of $P_1$, we proceed to evaluate $P_1$. To evaluate each individual $\mathbf{x}_u^i$ within $P_1$, LL search is conducted using an MOEA to identify the corresponding LL PS approximation $\{\mathbf{x}_l^{i*}\}$ (Algo.~\ref{algo:blmo_framework} Line \ref{line:LLsearch_causalcheck}). In case $\{\mathbf{x}_l^{i*}\}$ is empty, such a solution is assigned a $\infty$ value at UL. The special cases of VAA require additional search components, which will be discussed later in Algos.~\ref{algo:causal_LL_search}-\ref{algo:causal_UL_search}.

The $\mathbf{x}_u$ of the non-dominated~(ND) solutions in $P_1$ and the $\{\mathbf{x}_l^{*}\}$ corresponding to each of them are paired and stored in an archive (Algo.~\ref{algo:blmo_framework} Line~\ref{framework:archive1}). All solutions that went through LL search are accumulated into the archive for training/updating the PSP model (Algo.~\ref{algo:blmo_framework} Line \ref{framework:nn_update}). 
As indicated in the previous section, the helper variable $r$ is also generated and appended to $\mathbf{x}_u$ in the archive. After the first population $P_1$ has been evaluated through LL search, UL offspring population is generated by using evolutionary operators~(Algo.~\ref{algo:blmo_framework}, Line \ref{framework:child_generation}). Since the PSP model has been trained in the previous step, it can be used to supplement LL search for evaluation of the UL new offspring. This can be done by initializing the LL population using the predicted PS~(Algo.~\ref{algo:generator_assisted_LLsearch}) for the LL search. Here, we introduce an NN update parameter $\gamma$ to explore the potential of the predictor, and  a parameter $ds$ for training data size threshold. Without sufficient training data, the accuracy of the model may not be good enough to reliably skip LL search. Therefore, LL search is conducted as usual, until training data size meets $ds$~(Algo.~\ref{algo:blmo_framework}, Line~\ref{line:gamma}). Thereafter, the LL search is only run every $\gamma$ generations, initialized with PSP. The LL search is skipped in the intermediate generations entirely~(Algo.~\ref{algo:blmo_framework}, Line~\ref{framework:LL_skip}). In this way, the LL search can be forgone for most of the generations to save on evaluations. The training data for NN are only accumulated from solutions for which LL search is conducted to ensure the accuracy of the evaluations in the dataset~(Algo.~\ref{algo:blmo_framework}, Line~\ref{framework:selective_update}). When the number of data points in the archive exceeds $ds$, the most recent $ds$ data points are used for building the model, to keep the model more accurate in the current region of search.
After each UL child has received its corresponding LL solutions $\{\mathbf{c}_l^{i*} \}$ and evaluated at UL~(Algo.~\ref{algo:blmo_framework}, Line~\ref{framework:eval_child}), UL ND archive is first updated. Then parent and child populations are combined and sorted according to feasibility first~(FF), non-dominance~(ND) and crowding distance~(CD)~(Algo.~\ref{algo:blmo_framework}, Line~\ref{framework:sorting}). The next generation UL population is selected from combined sorted population using environmental selection mechanism outlined in Algo.~\ref{algo:environment_selection}. This evolution continues until termination criterion is met, and the archive of ND solutions obtained is returned.

\subsection{PSP and PSP-assisted LL search}
The role of PSP is to make LL search efficient by either replacing it altogether with predicted PS, or providing LL search a better starting population than random initialization. For LL PS prediction, when a new $\mathbf{x}_u$ is given to PSP, a list of linearly spaced helper value $r$~($r\in[0, 1]$) is first generated (Algo.~\ref{algo:nn_init} Line~\ref{nn_init_r}). Then $\mathbf{x}_u$ is replicated to achieve the same number~($n^l$) of solutions that exist in $r$ (Algo.~\ref{algo:nn_init} Line~\ref{nn_expand}). The collated $\mathbf{x}_u$ and $r$ are provided as an input for PSP ($\phi$) (Algo.~\ref{algo:nn_init} Line~\ref{nn_predict}). The output of $\phi$ forms candidate solutions for the LL PS or initial solutions for LL search~(Algo.~\ref{algo:nn_init} Line~\ref{nn_return}).

\begin{algorithm}[!ht] \footnotesize
\caption{Generate LL solutions by PSP} 
\begin{flushleft}
\textbf{Input:} Population size $n^l$; upper level solution $\mathbf{x}_u$; PSP model $\phi$ \\
\textbf{Output:} A set of $\{\mathbf{x}_l\}$ solutions $P_1^l$
\end{flushleft}
\begin{algorithmic}[1]
\STATE Generate $n^l$ linearly spaced values within $[0, 1]$ in order as $R$ \label{nn_init_r}
\STATE Replicate $\mathbf{x}_u$ to size $n^l$ as $X$; normalize $X$ using UL variable bounds \label{nn_expand}
\STATE Combine $[X, R]$ to form input of network $\phi$ 
\STATE Generate population $P_1^l = \{\mathbf{x}_l^1,..., \mathbf{x}_l^{n^l}\}$ by using $\phi([X, R])$ \label{nn_predict}
\STATE De-normalize $P_1^l$ using UL variable bounds
\STATE Return $P_1^l$ \label{nn_return}
\end{algorithmic}
\label{algo:nn_init}
\end{algorithm}

The above generated solutions are evaluated using the LL objective/constraint functions. Then, for the generations where PS approximation is directly used, the ND of the above solutions are returned back to the UL. For the generations where LL search is invoked~(i.e., every $\gamma$ generations; Algo.~\ref{algo:blmo_framework}, Line~\ref{line:gamma}), the ND solutions are used to seed the first LL population~(Algo.~\ref{algo:generator_assisted_LLsearch}, Line~\ref{line:nd_init}).

\begin{algorithm}[!ht] \footnotesize
\caption{PSP assisted LL search} 
\begin{flushleft}
\textbf{Input:} UL solution $\mathbf{x}_u$, Population size $n^l$; PSP model $\phi$, Evolutionary parameters, LL association vector $v$\\
\textbf{Output:} Search result $L^*$ of LL problem
\end{flushleft}
\begin{algorithmic}[1]
\STATE Use Algo.~\ref{algo:nn_init} to generate a population $P^l=\{\mathbf{x}_l\}$ 
\STATE Evaluate $P^{l}$ on LL objectives
\STATE Identify ND solutions in $P^{l}$ as $P_{ND}^{l}$ 
\STATE Obtain $L^*$ via Algo.~\ref{algo:causal_LL_search} with $P_{ND}^{l}$ inserted into the initial population ~\label{line:nd_init}
\STATE Return $L^*$ 
\end{algorithmic}
\label{algo:generator_assisted_LLsearch}
\end{algorithm}

For the LL searches in the first UL generation, the initial population is randomly generated, while subsequently it comes partially from predicted LL ND solutions. If ND solution size is smaller than pre-defined population size $n^l$,  then the rest of the solutions are filled in with randomly generated solutions (Algo.~\ref{algo:causal_LL_search} Line~\ref{line:fill_random}). The intent behind this is that if the initial solutions generated by PSP model $\phi$ are clustered in a small region, these randomly introduced candidate solutions can help maintain diversity to a certain extent. Inferred from vector $v$, an LL solution can be divided into $(\mathbf{x}_{li}^U, \mathbf{x}_{li}^L)$, where $\mathbf{x}_{li}^U$ refers to LL variables that only appear in UL objective. If there is no VAA in the problem~(which is most often the case), then $\mathbf{x}_{li}^U$ is empty. LL search discussed above is only applied to $\mathbf{x}_{li}^L$ (Algo.~\ref{algo:causal_LL_search} Line.~\ref{line:partial_search}). This brings benefits of smaller search space and potentially quicker convergence speed to LL search, for the cases where $\mathbf{x}_{li}^U$ is non-empty.

After first population $P_1^l$ is formed and evaluated, evolutionary operators are used to generate child population $C_g$ (Algo.~\ref{algo:causal_LL_search}, Line~\ref{line:generate_llchild}). Environmental selection (Algo.~\ref{algo:environment_selection}) is then invoked to select the surviving population. This iterates until stopping condition is met. Then ND solutions from the last population are selected (Algo.~\ref{algo:causal_LL_search}, Line~\ref{line:select_LLND}).
With $\mathbf{x}_{l}^L$ variables determined in $L$, we still need to identify the values for $\mathbf{x}_{l}^U$ by calling Algo~\ref{algo:causal_UL_search}, if VAA is present~(i.e., $\mathbf{x}_{li}^U$ is non-empty), as discussed next.

\begin{algorithm}[!ht] \footnotesize
\caption{MOEA LL search considering VAA}
\begin{flushleft}
\textbf{Input:} LL variable association vector $v$; Initial population $P_1^{l}$; Evolutionary parameters; Population size $n^l$; Stopping condition \\
\textbf{Output}: LL search result $L^*$ 
\end{flushleft}
\begin{algorithmic}[1]
\STATE Use $v$ to separate every solution $\mathbf{x}_{li}$ into $(\mathbf{x}_{li}^{U}, \mathbf{x}_{li}^{L})$; \label{line:partial_search}
\\
\textbf{// Search on $\mathbf{x}_{li}^L$ only; $\mathbf{x}_{li}^{U}$ is set to a random value //}
\IF{Initial population $P_1^{l}$ size is smaller than $n^l$ }
\STATE Add randomly generated LL solutions to bring $P_1^{l}$ size to $n^l$\label{line:fill_random}
\ENDIF
\STATE Set generation counter $g = 2$
\WHILE{stopping condition is not met}
\STATE Apply evolutionary operators to $\mathbf{x}_l^L$ part to generate child population. $C_g$. Discard previously evaluated/duplicate solutions from $C_g$\label{line:generate_llchild}
\STATE Evaluate $C_g$ and Sort $P_{g-1} \cup C_g$
\STATE Select new parent population $P_g$ using Algo.~\ref{algo:environment_selection} \label{line:env-select2}
\STATE Increment generation counter $g = g + 1$
\ENDWHILE
\STATE Select the ND front~($L$) of the last generation population \label{line:select_LLND}\\
\textbf{// Additional search on $\mathbf{x}_{li}^U$; $\mathbf{x}_{li}^{L}$ set to their optimized values//}
\IF{$\mathbf{x}_{li}^U$ is non-empty}
\STATE Call Algo.~\ref{algo:causal_UL_search} for each solution in $L$ to improve its $\mathbf{x}_{li}^{U}$ values
\ENDIF
\STATE Assign updated $L$ to $L^*$
\STATE Return $L^*$
\end{algorithmic}
\label{algo:causal_LL_search}
\end{algorithm}

\subsection{Additional UL search in case of VAA}

This module determines the value of $\mathbf{x}_l^U$ part of the search result $L$ from Algo~\ref{algo:causal_LL_search}. For each solution $\mathbf{x}_{li} = (\mathbf{x}_{li}^U, \mathbf{x}_{li}^{L*})$ in $L$, to determine values for the part $\mathbf{x}_{li}^U$, MOEA search is conducted to optimize the UL objectives with $\mathbf{x}_{li}^U$ as variables, while $\mathbf{x}_{li}^{L*}$ stay fixed. However, running a standard MOEA for every single $\mathbf{x}_{li}$ will require exorbitant FE consumption. To expedite this, we utilize solution transfer in this MOEA search. As shown in Algo.~\ref{algo:causal_UL_search} Lines~\ref{line:transfer_borrow}-\ref{line:transfer_insert}, once there are one or more solutions available in $L^*$, then when new search to be started, solution transfer is using solutions from $L^*$. For this, the solution $\mathbf{x}_{lj}^{L*}$ that has the smallest Euclidean distance from $\mathbf{x}_{li}^L*$~($i\neq j$) is identified. Then, $\mathbf{x}_{lj}^{U*}$ corresponding to this solution is inserted into the initial population for the search corresponding to this $\mathbf{x}_{li}^L$. Once $\mathbf{x}_{li}^{U*}$ is determined for each $\mathbf{x}_{li}^{L*}$, the updated $L^*$ is returned. 

\begin{algorithm}[!ht] \footnotesize
\caption{Additional UL search considering VAA}
\begin{flushleft}
\textbf{Input:} Evolutionary parameters, LL search result $L$, LL variable association vector $v$\\
\textbf{Output:} Updated LL search result $L^*$
\end{flushleft}
\begin{algorithmic}[1]
\STATE For each $\mathbf{x}_{li}$ in $L$, separate its variables into  $(\mathbf{x}_{li}^U, \mathbf{x}_{li}^{L*})$ using $v$
\STATE set  $i=1$,$n=$ size of $L$ and $L^* = \emptyset$
\WHILE{$i \leq n$}
\IF{$i = 1$}
\STATE Using UL objective to search on $\mathbf{x}_{li}^{U}$ with $\mathbf{x}_{li}^{L*}$ fixed
\STATE Add resulting $\mathbf{x}_{li}^* = (\mathbf{x}_{li}^{U*},  \mathbf{x}_{li}^{L*})$ to $L^*$; $i=i+1$
\ELSE
\IF{There are UL feasible solutions in $L^*$ }
\STATE Find UL feasible solution $\mathbf{x}_{lj}^{L*}$ in $L^*$ with closest Euclidean distance to $\mathbf{x}_{li}^{L}$; $i\neq j$\label{line:transfer_borrow}
\STATE Insert corresponding $\mathbf{x}_{lj}^{U*}$ to initial search population \label{line:transfer_insert}
\STATE Run MOEA on $\mathbf{x}_{li}^{U}$ until stopping condition is met
\STATE Add search result $\mathbf{x}_{li}^{*} = (\mathbf{x}_{li}^{U*}, \mathbf{x}_{li}^{L*})$ to $L^*$; $i=i+1$
\ELSE
\STATE Using UL objective to search (MOEA with predefined evolutionary parameters) on $\mathbf{x}_{li}^{U}$ with $\mathbf{x}_{li}^{L*}$ fixed
\STATE Add resulting $\mathbf{x}_{li}^* = (\mathbf{x}_{li}^{U*}, \mathbf{x}_{li}^{L*})$ to $L^*$; $i=i+1$
\ENDIF
\ENDIF
\ENDWHILE
\STATE Return $L^*$ 
\end{algorithmic}
\label{algo:causal_UL_search}
\end{algorithm}

\subsection{Environmental Selection}
\label{dss}
In the evolutionary process of Algo.~\ref{algo:blmo_framework}~(Line~\ref{line:env-select1}), and Algo.~\ref{algo:generator_assisted_LLsearch}, (Line~\ref{line:env-select2}) new population is selected from the sorted combined parent and offspring population. In order to maintain diversity in the population, we adopt distance based subset selection~(DSS)~\cite{DSS2018} in this step when ND front size has more unique solutions than predefined population size. As shown in Algo.~\ref{algo:environment_selection}, Line~\ref{check_ndsize}, from sorted population, we first identify the number $N$ of unique ND solutions in $\{\mathbf{x}\}_g$. If $N$ is smaller than predefined population size, then new population is selected in order of ranking. If $N$ is larger than predefined population size, then DSS selection is applied to $\{\mathbf{x}\}_g$'s ND front solutions. DSS selects solutions iteratively by choosing the one with the maximum distance to already selected solutions. As in Eq.~\ref{dss}, $\mathbf{x}_i$ refers to candidate solution, $\mathbf{x}_j^s$ refers to solutions already selected, and $k$ is the number of selected solutions. In each iteration, the solution that has highest value of $d_{\mathbf{x}_i}$ is selected. 
\begin{equation}\footnotesize
    d_{\mathbf{x}_i} = min\{d(\mathbf{x}_i, \mathbf{x}_1^s)), d(\mathbf{x}_i, \mathbf{x}_2^s), ...d(\mathbf{x}_i, \mathbf{x}_j^s), ...d(\mathbf{x}_i, \mathbf{x}_k^s)\}
    \label{dss}
\end{equation}

\begin{algorithm}[!ht] \footnotesize
\caption{DSS based environmental selection} 
\begin{flushleft}
\textbf{Input:} Sorted population $\{\mathbf{x}\}_g$; Sorted population fitness $F_g$; Expected new population size $n$  \\
\textbf{Output:} New population $\{\mathbf{x}\}_{g+1}$ and its fitness $F_{g+1}$\\
\end{flushleft}
\begin{algorithmic}[1]
\STATE Check the number of unique solutions ($N$) in the ND front of $F_g$ \label{check_ndsize}
\IF{$N <= n$}
\STATE Select top $n$ unique solutions from the $\{\mathbf{x}\}_g$ as $\{\mathbf{x}\}_{g+1}$
\STATE Extract $\{\mathbf{x}\}_{g+1}$'s corresponding objective values from $F_g$ to form $F_{g+1}$
\ELSE
\STATE Select $n$ solutions $\{\mathbf{x}\}_g$ of ND set by applying DSS
\STATE Assign the above selected solutions to $\{\mathbf{x}\}_{g+1}$
\STATE Extract above $\{\mathbf{x}\}_{g+1}$'s corresponding objective values from $F_g$ and assign to $F_{g+1}$
\ENDIF
\STATE Return $\{\mathbf{x}\}_{g+1}$ and $F_{g+1}$
\end{algorithmic}
\label{algo:environment_selection}
\end{algorithm}

\subsection{Termination condition}
\label{sec:stopping_condition}
To terminate the algorithm, we adopt the condition proposed in \cite{Blank2020}, which  
attempts to detect the stability~(stagnation) of the evolving population based on measurements capturing convergence and diversity. Convergence is measured by the following normalized ideal ($z^*$) difference and nadir ($z^{nad}$) difference. 

\begin{equation}\footnotesize
    \Delta z_{t-1, t}^* = max_{i=1}^M \frac{z_i^*(t-1) - z_i^*(t)}{z_i^{nad}(t) - z_i^*(t)}
\end{equation}

\begin{equation}\footnotesize
    \Delta z_{t-1, t}^{nad} = max_{i=1}^M \frac{z_i^{nad}(t-1) - z_i^{nad}(t)}{z_i^{nad}(t) - z_i^*(t)}
\end{equation}
If $\Delta z_{t-1, t}^*$ and $\Delta z_{t-1, t}^{nad}$ do not change significantly in the past $\omega$ generations by comparing their maximum metric values within a threshold $\epsilon$, then the algorithm is deemed to have sufficiently converged. Here $t$ refers to a given generation during evolution. $M$ refers to the number of objectives.

The diversity is assessed using the following metric~(but as noted in \cite{Blank2020}, is not entirely independent of convergence). 

\begin{equation}\footnotesize
    \phi(t) = IGD(\Bar{P}^t(t-1), \Bar{P}^t(t))
\end{equation} 

\noindent $IGD$ refers to inverted generational distance~\cite{coello2005solving}. Suppose $0\leq t \leq \tau$, $\tau$ is current generation, the normalized $i^{th}$ objective value of $j^{th}$ point in generation $t$ with regard to $\tau$ generation is computed as

\begin{equation}\footnotesize
    \Bar{P}_i^{\tau, j}(t) = \frac{P_i^j(t) - z_i^*{(\tau)}}{z_i^{nad}(\tau) - z_i^*(\tau)}
\end{equation}

A sliding window $\omega$ is used to compute the metrics. If over $\omega$ generations, the maximum values of three metrics given below are no greater than a threshold $\epsilon$, the algorithm is terminated. 

\begin{align}\footnotesize
    & max(\Delta z_{\tau, \tau-1}^*, ..., \Delta z_{\tau-\omega, \tau -\omega -1}^*) \leq \epsilon \\
   &  max(\Delta z_{\tau, \tau-1}^{nad}, ..., \Delta z_{\tau-\omega, \tau -\omega -1}^{nad}) \leq \epsilon\\
    & max(\Delta \phi_{\tau, \tau-1}, ..., \Delta \phi_{\tau-\omega, \tau -\omega -1}) \leq \epsilon
\end{align}

Similar to IGD, it is also possible to monitor HV to formulate a termination condition, as done in some of the other works~\cite{deb2010efficient, sinha2017evolutionary, Conditional_wang2023, cai2022cooperative}. Given that the true PF is not known a priori while solving the problem, either of these can only measure whether the population has stabilized, without guaranteeing convergence to the true optimum. We choose IGD based stopping condition as the default in our algorithm, merely to be consistent with the use of IGD metric later for performance assessment. IGD is a more reliable metric for measuring the final performance for BLMOPs, especially where deceptive problems are involved. For deceptive problems, a higher HV value may not reflect true performance of an algorithm, since it may be an artifact of sub-optimal solutions at the LL. The UL solutions in such cases may appear to dominate the true PF and generate high HV value, contradicting its true performance. IGD metric on the other hand becomes worse when PF approximation is away from the true PF, irrespective of which side it is located, providing a more accurate measurement. We adopt above IGD based stopping condition as default in our studies. However, some exceptions lie in  comparison with state-of-the-art algorithms where we have used termination based on HV or/and number of evaluations to be consistent with that used in other works.

\section{Experiments and discussion}
\label{sec:experiment}

Ten problems are used in our experiments from study~\cite{deb2010efficient}, which are listed in Table~\ref{tab:problem_settings}. Five of them~(i.e. TP1, TP2, DS1D, DS2D, DS3D) exhibit some level of deceptiveness. Theoretical PF can be derived for all the test problems to analyze the performance of the algorithms. To generate $n$ points approximately uniformly distributed on the PF, we first over-sample by generating $2n$ approximately uniform points on PF, utilizing the equations relevant to the PF of the problem. Then we use distance based subset selection~\cite{DSS2018} to select $n$ solutions~($n=1025$ to ensure near uniformity) as the final PF. For calculating HV when comparing to state-of-the-art algorithms, reference point is set 1.1 times of the maximum objective values of the true PF, without normalization, to be consistent with their settings.

\begin{table}[!ht]\footnotesize
    \centering
    \caption{Test problems for experiments.  $D_u$ and $D_l$ refers to number of variables on UL and LL, respectively}
    \renewcommand\arraystretch{0.8}
    \begin{tabular}{l r r  l r r }
    \hline
    Prob. &  UL($D_u$) & LL($D_l$) & Prob. & UL ($D_u$) & LL ($D_l$)\\
    \hline
       TP1 & 1 & 2   &  DS4 & 1 & 9 \\ 
       TP2 & 1&  14   & DS5 & 1 & 9\\ 
       DS1 & 10 &  10 &  DS1D & 10 & 10\\
       DS2 & 10 &  10 &  DS2D &  10 & 10\\
       DS3 & 10 &  10 &  DS3D & 10 & 10\\
     \hline
    \end{tabular}
    \label{tab:problem_settings}
\end{table}

For the evolutionary search part of the proposed framework, differential evolution~(DE)~\cite{storn1997differential} and polynomial mutation~\cite{deb2002fast} are used for generating offspring population. The crossover rate is set to 1 and scaling factor is set to 0.5 for DE operator. As for PM, its mutation probability is set to $1/D$ ($D$ is the number of variables) and mutation index to 20. The first generation of PSP-BLEMO relies on MOEA for its LL search. This MOEA has fixed generation size set relative higher to 300 to ensure first training data has good quality. Stopping conditions as discussed previously are activated during the remainder of the search. 21 independent runs are conducted for each setting. Wilcoxon rank-sum significant test is used to draw conclusions regarding statistical significance. 

For the NN structure used in PSP-BLEMO, one hidden layer is used. The size of hidden layer nodes is set to two times of the size of input nodes or output nodes, whichever is higher. Two parameters of PSP-BLEMO, re-train gap $\gamma$ and training data size $ds$, are determined empirically through experiments, discussed shortly in more detail; however it is to be noted that with sufficient training datasize, algorithm performance is observed to be not too sensitive to $\gamma$. Matlab \texttt{feedforwardnet} tool is used to train the NN with all parameters at their default values. For example, learning rate is 0.1, loss function is mean square error between network output  and target output. Levenberg-Marquardt is the default training algorithm. NN training stops if validation error fails to decrease for 6~(default) iterations. The split ratio for training, validation and test is 70\%, 15\% and 15\%, respectively.

In the following subsections, we present three sets of experiments. In the first set, we empirically determine two key parameters of PSP-BLEMO, namely the training gap $\gamma$ and the data size threshold $ds$. After their values are fixed, we continue on to the second set of experiments, which compare the performance of PSP-BLEMO with six state-of-the-art algorithms, namely cG-BLEMO~\cite{Conditional_wang2023}, MOBEA-DPL~\cite{wang2022multi}, SMS-MOBO~\cite{mejia2022novel}, BLMOCC~\cite{cai2022cooperative}, mf-BLEAQ~\cite{sinha2017evolutionary}, H-BLEMO~\cite{deb2010efficient}. For these comparison studies, both UL and LL population sizes are fixed to 20, same as in the most recent study~\cite{Conditional_wang2023}. The parameter settings for SMS-MOBO, are consistent with \cite{Conditional_wang2023} too. As discussed in above stopping condition section, for these experiments, HV stopping condition~\cite{deb2010efficient} is used to be consistent with these methods. The thresholds for HV stopping condition are $1e-3$ over $\omega=10$ generations, same as compared methods in~\cite{Conditional_wang2023, cai2022cooperative,sinha2017evolutionary}. For the additional UL search, population size is set to 5 and first search generation size is set to 80. During the UL search assisted by transferred solutions (Algo.~\ref{algo:causal_UL_search}), HV stopping condition is used for consistency.  In the last set of experiments, we compare PSP-BLEMO with its variant where the training is done only once~(referred to as one-shot~(OS) method) in order to save on evaluations. It reveals some interesting observations are when larger population sizes~(proportional to number of variables) are used, unlike the previous two experiments.

\subsection{Empirical experiments on key parameters}

In this first experiment, we use empirical analysis to gauge the effect of $\gamma$ and $ds$. For $\gamma$, we investigate 5 different values, $5, 10, 15, 20$ and infinite~($Inf$). $Inf$ here means that once the data reaches $ds$ size, the model is trained and thereafter never updated. As for $ds$, we tested 5 values: $5e2, 1e3, 2e3$ and $5e3$. Test bed involves all 10 problems shown in Table~\ref{tab:problem_settings}. For each combination of $\gamma$ and $ds$, 21 runs are conducted for each problem. Both UL and LL stops using default IGD stopping condition, with thresholds ($\epsilon$) being 1e-2 over 5 generation window~($\omega$). The statistics on IGD median values and FE consumption are reported in the supplementary~(Section 1). Here, we include a summary of statistical significance tests in Table~\ref{tab:keyparameter_sigtest}. For each $\gamma$ setting, we run Wilcoxon rank sum tests~(0.05 significance level) between each data size setting to the others on each problem, and count how many times this setting shows significant better performance. For example, when $\gamma=5$ is fixed, run statistical tests on test problems between (1) $ds=5e2$ vs $ds=1e3$, (2) $ds=5e2$ vs $ds=2e3$, (3) $ds=5e2$ vs $ds=5e3$, resulting in 30 tests~(10 test problems $\times$ 3 pairwise comparisons). Setting $ds=5e2$ shows significant better, equivalent and worse performance in 0, 13 and 17 tests, respectively. 

\begin{table*}[!ht] \footnotesize
 \caption{Statistical analysis outcomes on key parameters. Pairwise comparisons on the left are for fixed $\gamma$ in each column, while those on the right are done for fixed $ds$ in each column. The numbers represent significantly better, equivalent and worse cases across all problems.}
    \centering
    \setlength{\tabcolsep}{3pt}
    \begin{tabular}{l| l l l l  l  |l | l l l l }
    \hline
    Settings &  $\gamma=5$ & $\gamma=10$ & $\gamma=15$ & $\gamma=20$ &  $\gamma=Inf$ &  Settings & $ds=5e2$ & $ds=1e3$ & $ds=2e3$ & $ds=5e3$\\
    \hline
    $ds=5e2$ & 0-13-17 & 0-10-20 & 0-10-20 & 1-5-25 & 0-1-29 & $\gamma=5$ & \textbf{25}-15-0 &  \textbf{19}-21-0 &  \textbf{15}-24-1 & 0-35-5 \\
    $ds=1e3$ & 4-16-10  & 5-16-9  & 7-13-10 & 7-13-10 & 9-2-18 & $\gamma=10$ & 15-20-5  & 11-25-4 & 7-33-0 & \textbf{5}-35-0 \\
    $ds=2e3$ & 12-15-3  & 11-16-3 & 13-14-3 & 13-13-4 & 20-7-3 & $\gamma=15$ & 13-21-6 & 10-25-5 & 7-32-1 & 1-37-2\\
    $ds=5e3$ & \textbf{15}-14-1 & \textbf{17}-13-0 & \textbf{17}-12-1 &  \textbf{17}-13-0 & \textbf{22}-8-0 & $\gamma=20$ & 4-20-16 
    &  4-26-10 & 1-30-9 & 3-36-1\\
     \cline{1-6}      
       \multicolumn{6}{r|}{Out of 30 pairwise tests}               &      $\gamma=Inf$ & 0-30-10 & 0-25-15 & 1-26-13 & 1-37-2\\
    \cline{7-11}
     \multicolumn{11}{r}{Out of 40 pairwise tests}               
    \end{tabular} 
    \label{tab:keyparameter_sigtest}
\end{table*}

The left half of Table~\ref{tab:keyparameter_sigtest} basically evaluates which data size $ds$ setting has most win cases when $\gamma$ setting is unchanged. It can be seen that $ds=5e3$ has highest win cases in all $\gamma$ settings. Therefore, in following experiments, we fix training data size $ds=5e3$. Following the same approach, the right half of Table~\ref{tab:keyparameter_sigtest} shows the number of win cases for each $\gamma$ setting when $ds$ is fixed. We can see that when $ds=5e3$, $\gamma=10$ has most win cases. Consequently $ds$ and $\gamma$ are set to these two values for further experiments. \\

\subsection{Comparison with state-of-the-art approaches}

The compared algorithms include the most recently proposed cG-BLEMO~\cite{Conditional_wang2023} along with MOBEA-DPL~\cite{wang2022multi}, SMS-MOBO~\cite{mejia2022novel}, BLMOCC~\cite{cai2022cooperative}, mf-BLEAQ~\cite{sinha2017evolutionary} and H-BLEMO~\cite{deb2010efficient}. For BLMOCC and H-BLEMO we inherit the reported results in their corresponding papers, due to unavailability of the codes. The key idea of PSP-BLEMO is to replace LL PS with predicted value, cG-BLEMO and mf-BLEAQ are the most closely related algorithms for comparison, as both involve modules predicting LL solutions. For mf-BLEAQ, the source code was not available and hence we implemented its LL PS predictor, i.e. set valued mapping~(SVM) and embedded it into PSP-BLEMO framework for comparisons.

\begin{table*}[!ht]\scriptsize
\caption{Performance comparison with state-of-the-art algorithms. Compared to PSP-BLEMO the symbols $\uparrow$,$\downarrow$,$\approx$  denote statistically significantly better, worse or equivalent performance, respectively, of the peer algorithm}
    \centering
        \tabcolsep 0.4mm
    \renewcommand\arraystretch{0.8}
     \begin{tabular}{l| r r   |r r | r r | r r| r r |r r | r r | r r }
        \hline
        \multirow{3}{*}{Prob.}  & \multicolumn{16}{c}{IGD} \\
        \cline{2-17}
        &   \multicolumn{2}{c}{PSP-BLEMO}  & \multicolumn{2}{|c}{cG-BLEMO} & \multicolumn{2}{|c}{cG-BLEMO (tr)} &  \multicolumn{2}{|c}{MOBEA-DPL} &  \multicolumn{2}{|c}{SMS-MOBO} &\multicolumn{2}{|c}{BLMOCC} & \multicolumn{2}{|c|}{H-BLEMO} & \multicolumn{2}{c}{SVM-BLEMO}\\
        \cline{2-17}
        &   mean & std & mean & std & mean & std &   mean & std &  mean & std   & mean & std & mean & std & mean & std\\
        \hline
DS1& 0.0448 	& 0.1511 	& 0.0135 	& 0.0055 	$\approx$ 	& 0.1626 	& 0.1580 	$\downarrow$ 	& 0.1129 	& 0.2482 	$\downarrow$ 	& \textbf{0.0056} 	& 0.0023 	$\uparrow$ 	& 0.0786 	& 0.0003 	$\approx$ 	& 0.1254 	& 0.1480 	$\approx$ 	& 35.4813 	& 22.3078 	$\downarrow$ 	\\
DS2& 0.0495 	& 0.0120 	& \textbf{0.0248} 	& 0.0160 	$\uparrow$ 	& 0.2203 	& 0.1842 	$\downarrow$	& 0.0397 	& 0.0187 	$\uparrow$ 	& 0.0648 	& 0.0457 	$\approx$ 	& 0.1087 	& 0.0059 	$\downarrow$ 	& 0.1782 	& 0.0516 	$\downarrow$ 	& 23.0514 	& 23.0274 	$\downarrow$ 	\\
DS3& 0.1463 	& 0.0586 	& 1.4684 	& 6.4086 	$\uparrow$ 	& 3.0589 	& 6.2685 	$\downarrow$ 	& 0.0970 	& 0.0427 	$\uparrow$ 	& \textbf{0.0745} 	& 0.0783 	$\uparrow$ 	& 0.1667 	& 0.0294 	$\approx$ 	& 0.2569 	& 0.0858 	$\downarrow$ 	& 37.9607 	& 17.0301 	$\downarrow$ 	\\
DS1D& \textbf{0.0117} 	& 0.0053 	& 0.4743 	& 0.9775 	$\downarrow$ 	& 1.3983 	& 1.1611 	$\downarrow$ 	& 14.3168 	& 15.6289 	$\downarrow$ 	& 0.7348 	& 0.6656 	$\downarrow$ 	& - 	& - 	& - 	& - 	& 3e4 	& 2e3 	$\downarrow$ 	\\
DS2D & \textbf{0.0422} 	& 0.0128 	& 0.7363 	& 2.1869 	$\approx$ 	& 1.8393 	& 2.1736 	$\downarrow$ 	& 14.4400 	& 14.8768 	$\downarrow$ 	& 10.8181 	& 5.4551 	$\downarrow$ 	& - 	& - 	& - 	& - 	& 3e3 	& 169.6525 	$\downarrow$ 	\\
DS3D& \textbf{0.1682} 	& 0.0645 	& 2.0446 	& 6.2895 	$\uparrow$ 	& 3.8883 	& 5.9489 	$\downarrow$ 	& 398.8608 	& 226.6369 	$\downarrow$ 	& 2.0868 	& 1.4613 	$\downarrow$ 	& - 	& - 	& - 	& - 	& 1e3 	& 875.1150 	$\downarrow$ 	\\
DS4& \textbf{0.0173} 	& 0.0009 	& 0.0317 	& 0.0070 	$\downarrow$ 	& 0.0292 	& 0.0095 	$\downarrow$ 	& 0.1003 	& 0.0218 	$\downarrow$ 	& 0.3683 	& 0.1671 	$\downarrow$ 	& 0.0689 	& 0.0035 	$\downarrow$ 	& 0.0772 	& 0.0027 	$\downarrow$ 	& 1.2404 	& 0.6522 	$\downarrow$ 	\\
DS5& \textbf{0.0150} 	& 0.0016 	& 0.0494 	& 0.0225 	$\downarrow$ 	& 0.0437 	& 0.0213 	$\downarrow$ 	& 0.1052 	& 0.0272 	$\downarrow$ 	& 0.8550 	& 0.1708 	$\downarrow$ 	& 0.0796 	& 0.0045 	$\downarrow$ 	& 0.0828 	& 0.0041 	$\downarrow$ 	& 1.1634 	& 0.5678 	$\downarrow$ 	\\
TP1& 0.0117 	& 0.0010 	& 0.0152 	& 0.0030 	$\downarrow$ 	& 0.0111 	& 0.0015 	$\approx$ 	& 0.0115 	& 0.0012 	$\approx$ 	& \textbf{0.0051} 	& 0.0024 	$\uparrow$ 	& 0.0523 	& 0.0001 	$\downarrow$ 	& 0.0514 	& 0.0368 	$\downarrow$ 	& 0.0161 	& 0.0148 	$\approx$ 	\\
TP2& 0.0114 	& 0.0029 	& 0.0084 	& 0.0028 	$\uparrow$ 	& 0.0075 	& 0.0016 	$\uparrow$  	& 0.0130 	& 0.0037 	$\approx$ 	& \textbf{0.0059} 	& 0.0017 	$\uparrow$ 	& 0.0249 	& 0.0039 	$\downarrow$ 	& 0.0039 	& 0.0487 	$\approx$ 	& 0.0082 	& 0.0026 	$\uparrow$ 	\\
        \hline
        \multirow{3}{*}{Prob.} & \multicolumn{16}{c}{HV} \\
        \cline{2-17}
       &  \multicolumn{2}{c}{PSP-BLEMO}  & \multicolumn{2}{|c}{cG-BLEMO} & \multicolumn{2}{|c}{cG-BLEMO} &  \multicolumn{2}{|c}{MOBEA-DPL} &    \multicolumn{2}{|c}{SMS-MOBO} &  \multicolumn{2}{|c}{BLMOCC} & \multicolumn{2}{|c|}{H-BLEMO} &\multicolumn{2}{c}{SVM-BLEMO}\\
        \cline{2-17}
        &  mean & std & mean & std & mean & std &  mean & std & mean & std & mean & std& mean & std & mean & std\\
       \hline
DS1& 1.1379 	& 0.2037 	& 1.1805 	& 0.0107 	& 0.6702 	& 0.1557 	& 1.0460 	& 0.3356 	& 1.1963 	& 0.0032 	& 1.1520 	& 0.0027 	& 1.0890 	& 0.0152 	& 0.0000 	& 0.0000 	\\
DS2& 0.5278 	& 0.0187 	& 0.5723 	& 0.0227 	& 0.5058 	& 0.1784 	& 0.5508 	& 0.0263 	& 0.5117 	& 0.0621 	& 0.5176 	& 0.0204 	& 0.4894 	& 0.0036 	& 0.0000 	& 0.0000 	\\
DS3& 1.1706 	& 0.1189 	& 1.2747 	& 0.3004 	& 0.1777 	& 0.2412 	& 1.2845 	& 0.0866 	& 1.3551 	& 0.1083 	& 1.0320 	& 0.0336 	& 0.9518 	& 0.0376 	& 0.0000 	& 0.0000 	\\
DS1D& 1.1829 	& 0.0101 	& 1.0143 	& 0.5068 	& 0.2140 	& 0.2921 	& 235.3649 	& 450.8322 	& 0.4852 	& 0.4287 	& - 	& - 	& - 	& - 	& 7e6 	& 9e6 	\\
DS2D& 0.5394 	& 0.0201 	& 0.4884 	& 0.1675 	& 0.1374 	& 0.1984 	& 218.4854 	& 385.3930 	& 0.0000 	& 0.0000 	& - 	& - 	& - 	& - 	& 6e6 	& 5e5 	\\
DS3D& 1.1205 	& 0.1321 	& 24.7015 	& 75.1974 	& 0.1861 	& 0.3061 	& 1e5 	& 1e5 	& 0.0694 	& 0.1912 	& - 	& - 	& - 	& - 	& 9e5 	& 1e6 	\\
DS4& 1.3786 	& 0.0025 	& 1.3335 	& 0.0186 	& 0.6275 	& 0.0072 	& 1.1684 	& 0.0480 	& 0.6403 	& 0.1717 	& 1.1220 	& 0.0296 	& 1.0760 	& 0.0225 	& 0.0812 	& 0.1513 	\\
DS5& 1.2508 	& 0.0027 	& 1.1766 	& 0.0446 	& 0.6186 	& 0.0176 	& 1.0565 	& 0.0555 	& 0.1059 	& 0.0745 	& 0.9416 	& 0.0331 	& 0.9158 	& 0.0318 	& 0.0786 	& 0.1199 	\\
TP1& 0.5158 	& 0.0028 	& 0.5420 	& 0.0051 	& 0.3741 	& 0.0025  	& 0.5343 	& 0.0016 	& 0.5173 	& 0.0062 	& 0.3632 	& 0.0037 	& 0.3614 	& 0.0023 	& 0.5194 	& 0.0281 	\\
TP2& 0.2904 	& 0.0015 	& 0.2913 	& 0.0013 	& 0.5301 	& 0.0020 	& 0.2922 	& 0.0019 	& 0.2870 	& 0.0003 	& 0.2264 	& 0.0037 	& 0.2023 	& 0.0022 	& 0.2913 	& 0.0045 	\\
        \hline        
  \end{tabular} 
        \label{tab:compare_metric}
\end{table*}

In Table~\ref{tab:compare_metric}, the mean and standard deviation of HV and IGD values obtained by the six algorithms are shown. 
The symbols $\uparrow$, $\downarrow$ and $\approx$ show significantly better, worse or equivalent result, respectively, compared to PSP-BLEMO. For BLMOCC and H-BLEMO the statistical tests are conducted through confidence interval test~\cite{moore2007basic}. If two algorithms' 95$\%$ confident intervals are overlapping, then there is no statistical difference between two algorithms, otherwise, statistical difference is reported. For others, Wilcoxon ranksum test is used. Since our experiments include deceptive problems, HV performance can be misleading~(as discussed earlier in Section~\ref{sec:stopping_condition}). We therefore mainly focus on IGD performance, but still show HV results for completeness. With regards to cG-BLEMO, we observed certain trade-offs between function evaluations and performance, so we have reported two values. The second column in Table~\ref{tab:compare_metric} shows the performance of cG-BLEMO after its run is completed. The third column is also cG-BLEMO, but marked with FE truncated (tr). The metric values of cG-BLEMO are extracted when its total FE consumption exceeds total FE consumption of PSP-BLEMO for each run~(across 21 runs). The process is as follows. Both PSP-BLEMO and cG-BLEMO runs are sorted based on the FE consumed and then the metric values of cG-BLEMO are extracted when its total FE consumption exceeds the FE consumption of PSP-BLEMO for the corresponding run in the sorted order. This truncation allows us to see performance difference between PSP-BLEMO and cG-BLEMO for similar computational budgets.

In terms of mean IGD~(Table~\ref{tab:compare_metric}), PSP-BLEMO shows the lowest value for 5 out of 10 problems, while cG-BLEMO has the lowest for 1. In terms of significance test, cG-BLEMO performs better than PSP-BLEMO in 4 problems, and worse or equal in the other 6 problems. PSP-BLEMO shows better mean and standard deviation values especially in deceptive problems when compared with cG-BLEMO. This implies that performance variation of cG-BLEMO is much higher when dealing with deceptive problems. This can be explained as follows. cG-BLEMO inserts UL evaluations on LL solutions during LL search. If its LL prediction module initializes LL solutions that are not close enough to PS, UL objectives of such LL solutions may dominate the true UL PF. 
Consequently, UL search may be misguided by evolving around corresponding UL solutions which show inaccurate UL objective values; resulting in poor search performance. In this respect, mean and standard deviation reflect how well LL prediction maintains its performance. PSP-BLEMO shows relatively stable performance compared to cG-BLEMO.

\begin{table*}[!ht]\scriptsize
\caption{A comparison of FE consumed until termination with the state-of-the-art algorithms}
    \centering
        \tabcolsep 0.4mm
    \renewcommand\arraystretch{0.8}
    \begin{tabular}{l| r r r  |r r r|r r r |r r r | r r r| r r r|r r r}
    \hline
         \multirow{3}{*}{Prob.}& \multicolumn{21}{c}{UL FE} \\
        \cline{2-22}
         & \multicolumn{3}{c}{PSP-BLEMO} & 
         \multicolumn{3}{|c}{$\frac{\text{cG-BLEMO}}{\text{PSP-BLEMO}}$} & 
         \multicolumn{3}{|c}{$\frac{\text{MOBEA-DPL}}{\text{PSP-BLEMO}}$} &
         \multicolumn{3}{|c}{$\frac{\text{SMS-MOBO}}{\text{PSP-BLEMO}}$} &
         \multicolumn{3}{|c|}{$\frac{\text{BLMOCC}}{\text{PSP-BLEMO}}$} & 
         \multicolumn{3}{c|}{$\frac{\text{H-BLEMO}}{\text{PSP-BLEMO}}$}&  
         \multicolumn{3}{c}{$\frac{\text{SVM-BLEMO}}{\text{PSP-BLEMO}}$}\\
         \cline{2-22}
         &   Min & Med & Max & Min  & Med & Max & Min  & Med & Max &Min  & Med & Max&Min  & Med & Max&   Min & Med & Max &   Min & Med & Max\\
        \hline
DS1& 52,608 	& 69,793 	& 97,934 	& 5.78 	& 7.85 	& 9.85 	& 23.82 	& 20.13 	& 25.91 	& 6.10 	& 4.60 	& 3.28 	& 0.68 	& 0.54 	& 0.44 	& 1.66 	& 1.32 	& 1.10 	& 0.04 	& 0.05 	& 0.06 	\\
DS2& 48,373 	& 61,406 	& 79,380 	& 7.01 	& 9.52 	& 12.25 	& 22.32 	& 28.05 	& 36.27 	& 3.33 	& 2.63 	& 2.03 	& 0.96 	& 0.84 	& 0.73 	& 2.18 	& 1.90 	& 1.74 	& 0.05 	& 0.05 	& 0.10 	\\
DS3& 43,496 	& 63,839 	& 105,519 	& 0.27 	& 7.38 	& 7.63 	& 8.24 	& 9.39 	& 7.71 	& 7.41 	& 5.05 	& 3.06 	& 1.15 	& 0.82 	& 0.52 	& 2.59 	& 1.86 	& 1.19 	& 0.02 	& 0.03 	& 0.08 	\\
DS1D& 46,362 	& 70,945 	& 102,574 	& 1.12 	& 10.02 	& 11.09 	& 5.39 	& 6.46 	& 12.45 	& 1.74 	& 1.14 	& 0.79 	& - 	& - 	& - 	& - 	& - 	& - 	& 0.27 	& 0.34 	& 0.51 	\\
DS2D& 41,275 	& 59,365 	& 84,293 	& 2.59 	& 15.57 	& 18.67 	& 5.96 	& 9.30 	& 14.28 	& 2.00 	& 1.39 	& 0.98 	& - 	& - 	& - 	& - 	& - 	& - 	& 0.26 	& 0.38 	& 0.61 	\\
DS3D& 41,911 	& 58,911 	& 112,801 	& 6.36 	& 8.53 	& 14.43 	& 0.38 	& 0.39 	& 0.32 	& 1.97 	& 1.40 	& 0.73 	& - 	& - 	& - 	& - 	& - 	& - 	& 0.02 	& 0.32 	& 0.43 	\\
DS4& 1,532,766 	& 1,616,702 	& 1,701,643 	& 0.04 	& 0.11 	& 0.39 	& 0.05 	& 0.08 	& 0.34 	& 0.03 	& 0.02 	& 0.02 	& 0.01 	& 0.02 	& 0.02 	& 0.02 	& 0.03 	& 0.03 	& 0.02 	& 0.18 	& 1.10 	\\
DS5& 1,562,030 	& 1,620,794 	& 1,703,230 	& 0.02 	& 0.08 	& 0.17 	& 0.05 	& 0.08 	& 0.37 	& 0.03 	& 0.02 	& 0.02 	& 0.02 	& 0.02 	& 0.03 	& 0.03 	& 0.03 	& 0.04 	& 0.02 	& 0.10 	& 0.54 	\\
TP1& 9,122 	& 10,974 	& 14,668 	& 12.81 	& 39.58 	& 59.98 	& 12.48 	& 12.78 	& 15.16 	& 9.03 	& 7.51 	& 5.62 	& 0.94 	& 0.85 	& 0.70 	& 1.36 	& 1.29 	& 1.10 	& 1.23 	& 2.25 	& 2.67 	\\
TP2& 8,348 	& 11,533 	& 15,569 	& 11.07 	& 23.49 	& 40.92 	& 24.65 	& 31.13 	& 27.90 	& 9.58 	& 6.94 	& 5.14 	& 1.30 	& 1.03 	& 0.85 	& 2.07 	& 1.62 	& 1.38 	& 1.51 	& 1.92 	& 3.21 	\\
        \hline
         \multirow{3}{*}{Prob.}& \multicolumn{21}{c}{LL FE} \\
        \cline{2-22}
         & \multicolumn{3}{c}{PSP-BLEMO} & 
         \multicolumn{3}{|c}{$\frac{\text{cG-BLEMO}}{\text{PSP-BLEMO}}$} & 
         \multicolumn{3}{|c}{$\frac{\text{MOBEA-DPL}}{\text{PSP-BLEMO}}$} & 
         \multicolumn{3}{|c}{$\frac{\text{SMS-MOBO}}{\text{PSP-BLEMO}}$} &
         \multicolumn{3}{|c|}{$\frac{\text{BLMOCC}}{\text{PSP-BLEMO}}$} & 
         \multicolumn{3}{c|}{$\frac{\text{H-BLEMO}}{\text{PSP-BLEMO}}$}&  
         \multicolumn{3}{c}{$\frac{\text{SVM-BLEMO}}{\text{PSP-BLEMO}}$}\\
         \cline{2-22}
         &   Min & Med & Max & Min  & Med & Max & Min  & Med & Max &Min  & Med & Max&Min  & Med & Max&   Min & Med & Max &   Min & Med & Max\\
        \hline
DS1& 635,944 	& 741,588 	& 886,319 	& 2.29 	& 3.52 	& 5.75 	& 14.17 	& 13.47 	& 16.24 	& 80.21 	& 68.78 	& 57.55 	& 1.37 	& 1.38 	& 1.34 	& 4.43 	& 4.62 	& 4.32 	& 0.44 	& 0.71 	& 1.02 	\\
DS2& 907,850 	& 1,120,954 	& 1,438,517 	& 2.89 	& 3.94 	& 5.85 	& 8.58 	& 13.05 	& 12.99 	& 28.23 	& 22.87 	& 17.82 	& 1.60 	& 1.49 	& 1.45 	& 4.94 	& 4.19 	& 3.80 	& 0.30 	& 0.32 	& 0.63 	\\
DS3& 1,075,021 	& 1,260,718 	& 1,579,870 	& 0.29 	& 5.97 	& 6.86 	& 9.38 	& 10.84 	& 11.08 	& 47.68 	& 40.66 	& 32.45 	& 1.20 	& 1.22 	& 1.16 	& 3.69 	& 3.75 	& 3.33 	& 0.21 	& 0.46 	& 1.23 	\\
DS1D& 640,250 	& 742,979 	& 898,860 	& 0.57 	& 6.83 	& 9.08 	& 4.04 	& 6.25 	& 14.03 	& 12.49 	& 10.77 	& 8.90 	& - 	& - 	& - 	& - 	& - 	& - 	& 3.48 	& 5.56 	& 10.48 	\\
DS2D& 810,721 	& 1,107,812 	& 1,680,088 	& 0.66 	& 6.02 	& 7.07 	& 2.25 	& 3.91 	& 4.74 	& 10.06 	& 7.36 	& 4.86 	& - 	& - 	& - 	& - 	& - 	& - 	& 1.84 	& 3.14 	& 5.69 	\\
DS3D& 1,010,271 	& 1,262,412 	& 1,634,291 	& 3.94 	& 5.66 	& 14.55 	& 0.78 	& 0.70 	& 0.74 	& 8.07 	& 6.46 	& 4.99 	& - 	& - 	& - 	& - 	& - 	& - 	& 0.22 	& 3.06 	& 6.87 	\\
DS4& 253,830 	& 265,555 	& 302,347 	& 0.48 	& 1.38 	& 4.30 	& 1.25 	& 2.01 	& 7.68 	& 15.76 	& 15.06 	& 13.23 	& 2.79 	& 2.83 	& 2.71 	& 5.34 	& 5.41 	& 5.54 	& 5.28 	& 21.16 	& 61.26 	\\
DS5& 253,103 	& 265,317 	& 301,466 	& 0.29 	& 0.98 	& 1.95 	& 1.28 	& 2.02 	& 8.25 	& 15.80 	& 15.07 	& 13.27 	& 3.21 	& 3.40 	& 3.52 	& 6.59 	& 6.75 	& 7.29 	& 5.27 	& 15.09 	& 37.55 	\\
TP1& 346,016 	& 368\,703 	& 403,948 	& 3.75 	& 14.80 	& 27.30 	& 4.20 	& 4.89 	& 7.06 	& 37.86 	& 35.53 	& 32.43 	& 1.20 	& 1.21 	& 1.21 	& 1.61 	& 1.71 	& 1.73 	& 0.49 	& 0.64 	& 1.07 	\\
TP2& 216,097 	& 228,868 	& 241,237 	& 1.47 	& 4.94 	& 9.37 	& 7.09 	& 11.58 	& 13.44 	& 118.10 	& 111.51 	& 105.79 	& 0.59 	& 0.67 	& 0.81 	& 1.22 	& 1.40 	& 1.48 	& 0.76 	& 0.92 	& 1.51 	\\
        \hline
        \end{tabular} 
        \label{tab:compare_FE}
\end{table*}

Next, we look into FE consumption of all algorithms, reported in Table~\ref{tab:compare_FE}. For brevity, raw values are reported for PSP-BLEMO, while for the rest of the algorithms, ratio of their FE to that of PSP-BLEMO is reported. It can be seen that cG-BLEMO uses around 7 to 40 times more FE than PSP-BLEMO at the UL. Likewise, on the LL, cG-BLEMO uses around 3 to 15 times more FEs than PSP-BLEMO. In previous IGD metric based performance, cG-BLEMO showed better statistical performance in 4 out of 10 cases, but it also consumes much higher resource. To have one of these two metrics (IGD or FE) on the same level, we report cG-BLEMO IGD performance when its FE consumption exceeds PSP-BLEMO's in column 3 Table~\ref{tab:compare_metric}.  Statistical analysis shows that PSP-BLEMO performs significantly better than cG-BLEMO in 8 out of 10 cases, suggesting the improved performance of PSP-BLEMO for lower evaluation budgets. To note that  DS4 and DS5 are two special cases, where additional UL search as suggested in Algo.~\ref{algo:causal_UL_search} are invoked. It can be observed that PSP-BLEMO consumes much more FEs for such problems especially on the UL. However, even when considering similar FEs, its performance still remain competitive to cG-BLEMO. In supplementary~(Section~II), we also present a comparison of PSP-BLEMO with cG-BLEMO for deceptive versions of DS4 and DS5~(formulated in \cite{bing2024BL}) to highlight the advantage of PSP-BLEMO for such problems. Additionally, in the supplementary~(Section III), we report FE comparisons based on when these algorithms achieve the same level of IGD as BLMOCC~\cite{Conditional_wang2023}. This method for comparison was adopted in \cite{Conditional_wang2023}, hence we report them for completeness. PSP-BLEMO is observed to be competitive in these comparisons, consistent with the results above.

\begin{figure*}[!ht]
\centering
    \subfloat[DS1]
    {\includegraphics[width=0.25\textwidth]{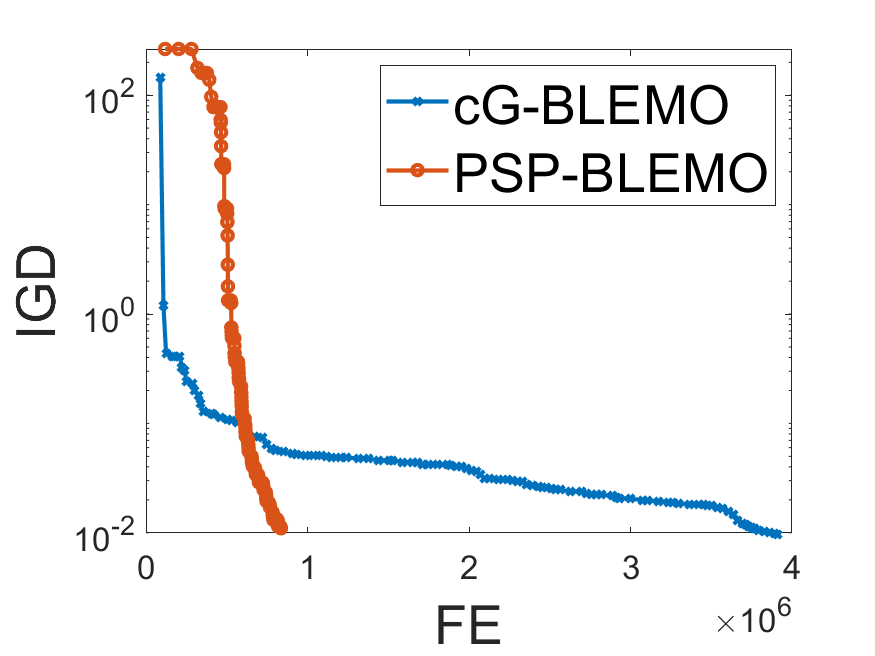}
        \label{fig:converge_ds1}}
    \subfloat[DS2]
    {\includegraphics[width=0.25\textwidth]{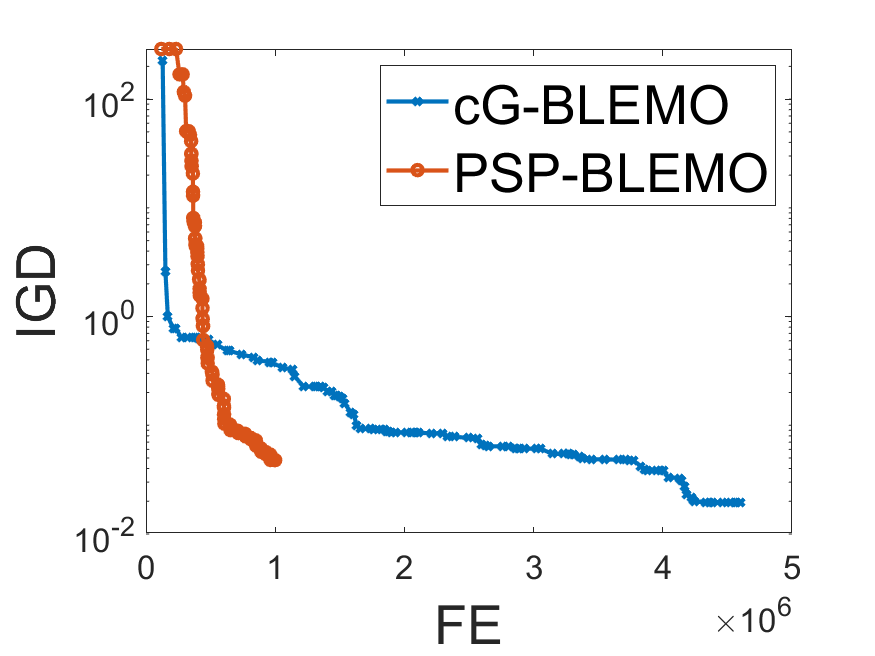}
    \label{fig:converge_ds2}}
\subfloat[DS3]{
\includegraphics[width=0.25\textwidth]{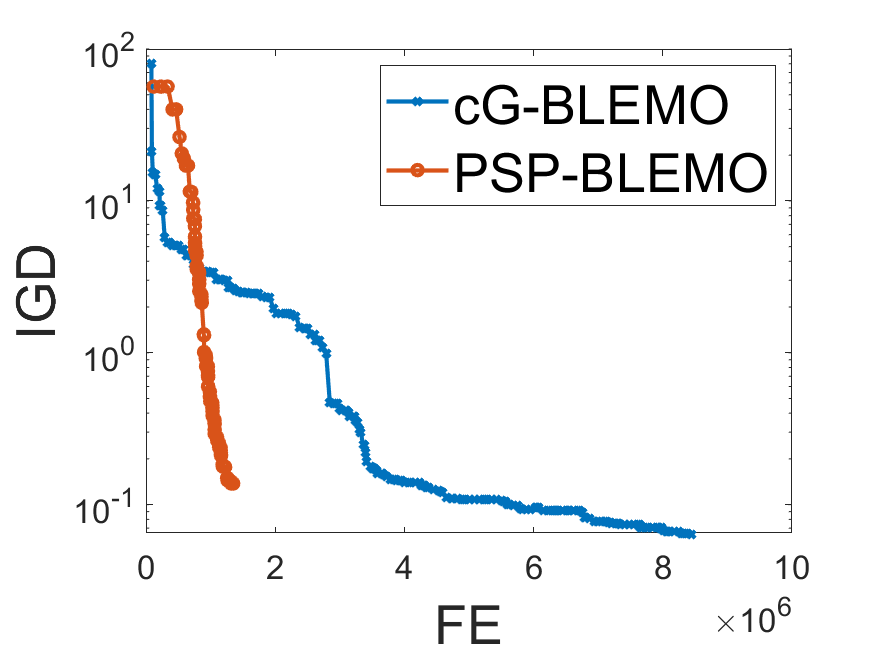}
    \label{fig:converge_ds3}}
\subfloat[DS1D]{
\includegraphics[width=0.25\textwidth]{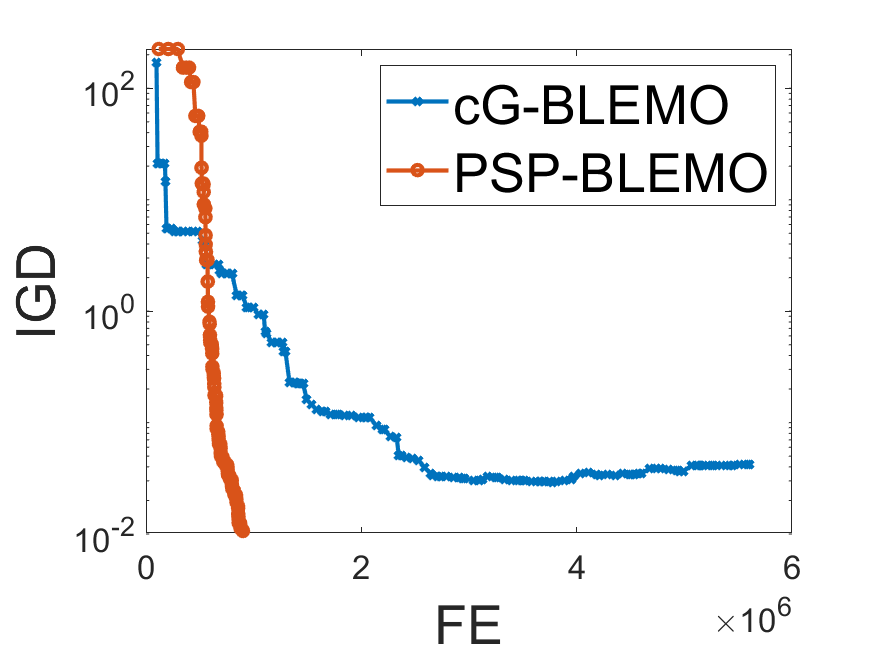}
    \label{fig:converge_ds1d}} \\
\subfloat[DS2D]{
\includegraphics[width=0.25\textwidth]{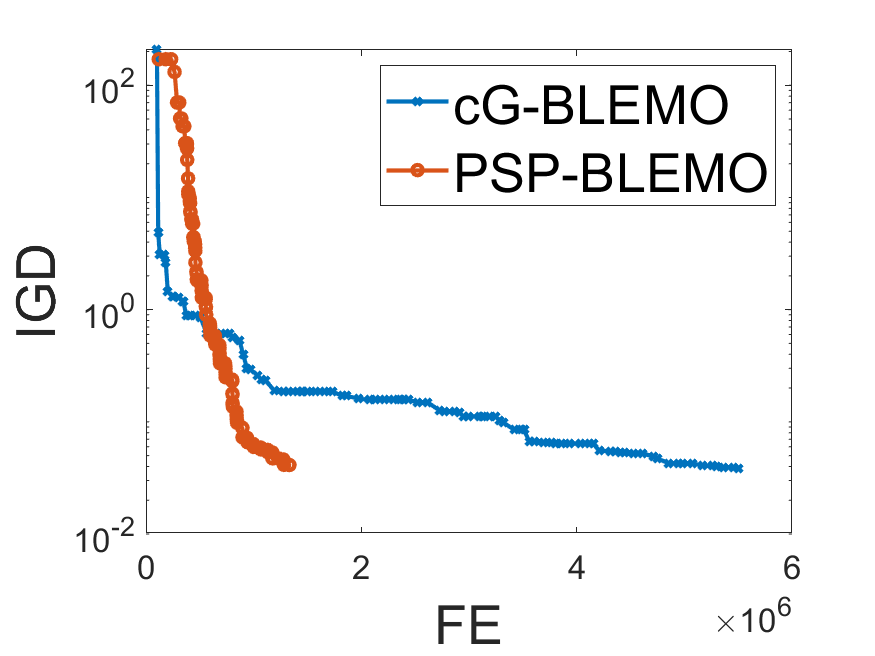}
    \label{fig:converge_ds2d}}
\subfloat[DS3D]{
\includegraphics[width=0.25\textwidth]{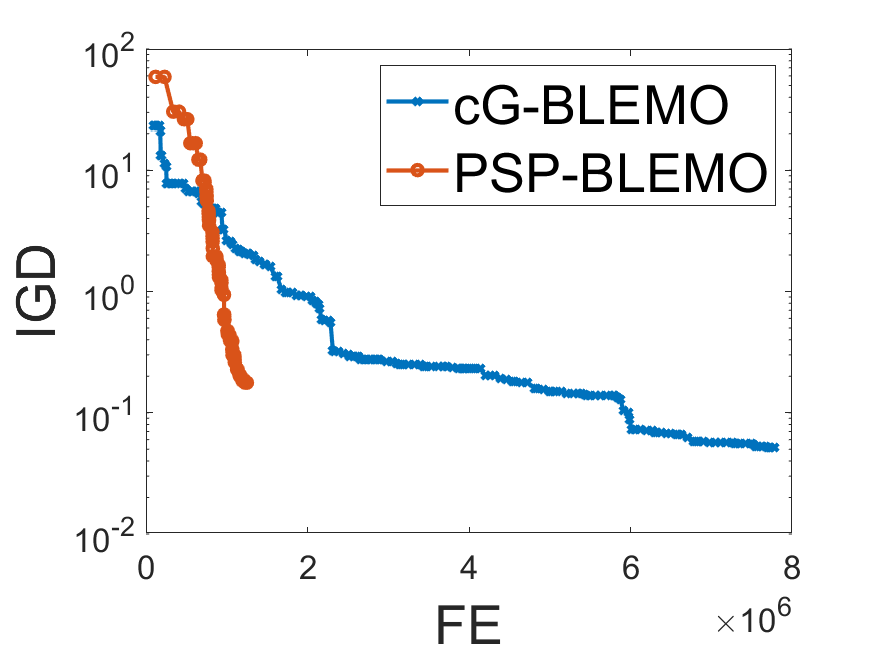}
    \label{fig:converge_ds3d}}
\subfloat[TP1]{
\includegraphics[width=0.25\textwidth]{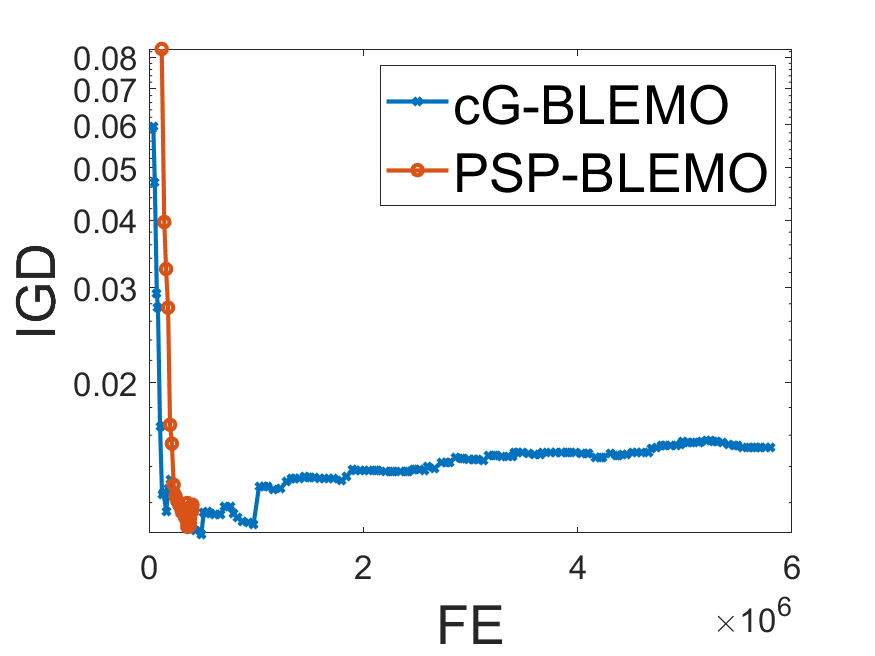}
    \label{fig:converge_tp1}}
\subfloat[TP2]{
\includegraphics[width=0.25\textwidth]{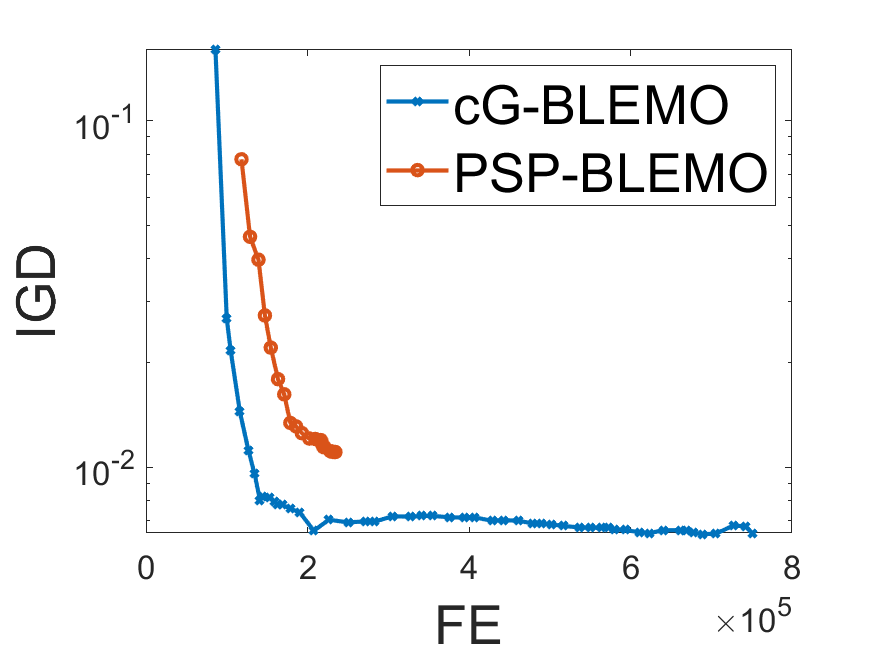}
    \label{fig:converge_tp2}} \\
\subfloat[DS4]{
\includegraphics[width=0.25\textwidth]{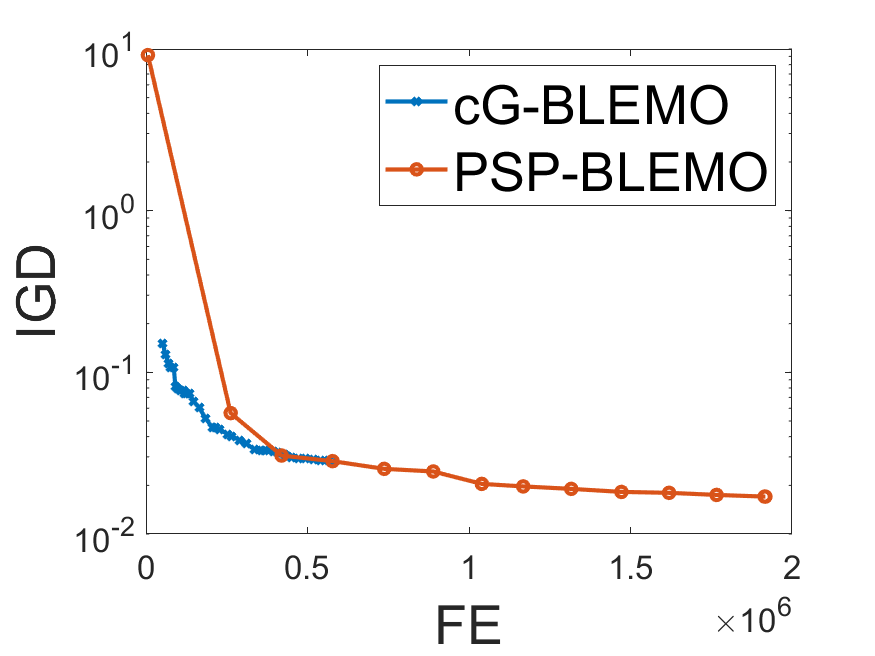}
    \label{fig:converge_tp1}}
\subfloat[DS5]{
\includegraphics[width=0.25\textwidth]{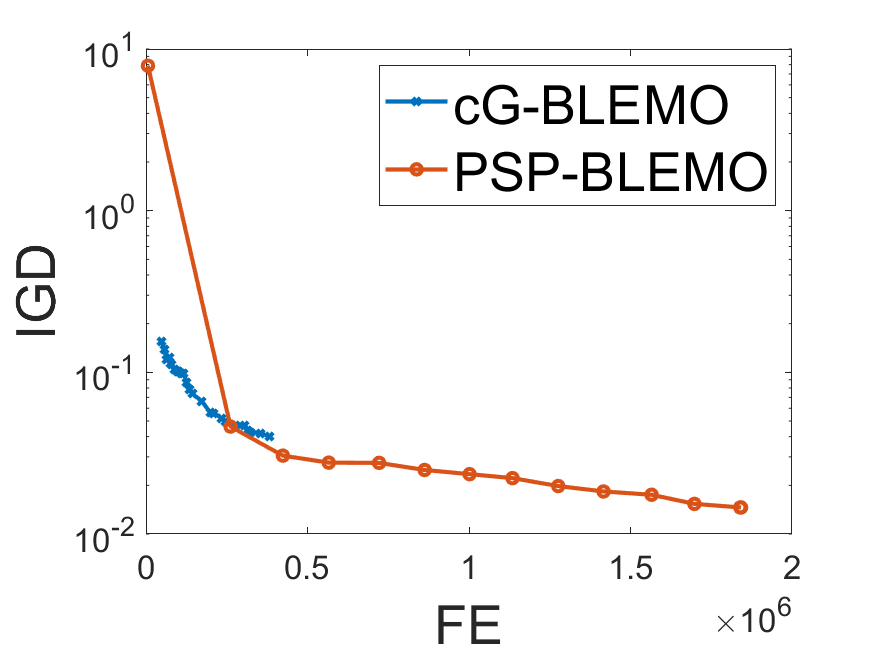}
    \label{fig:converge_tp2}}
\caption{Median result convergence plots. IGD values are shown in log scale.}
    \label{fig:convergence2}
\end{figure*}

The above performance comparison between PSP-BLEMO and cG-BLEMO is also visually reflected in convergence plots of median IGD runs shown in Fig~\ref{fig:convergence2}. Typically, PSP-BLEMO uses much fewer FEs to converge, at a faster rate than cG-BLEMO. Besides FE and IGD, there is also additional information implicit in the plots. For TP1 and DS1D, IGD values of cG-BLEMO decreases first and then rises. This is reflection of the algorithm performance affected negatively by the deceptive problems. In contrast, such case do not occur for PSP-BLEMO. For problems DS4 and DS5~(the cases with VAA), it can be observed that there is suddenly drop in IGD value in the first generation for PSP-BLEMO. It shows the effect of UL additional search to improve the IGD. In Table~\ref{tab:runtime}, we compare the average runtime of two algorithms. Consistent with FE consumption, it can be seen that PSP-BLEMO uses much less time to converge in 8 out of 10 problems. For DS4 and DS5, the two algorithms use similar time.

\begin{table}[!ht]\footnotesize
\centering
\caption{Comparison of average run time (unit: seconds)}
\renewcommand\arraystretch{0.8}
\setlength\tabcolsep{1mm} 
    \begin{tabular}{l| r r  | l| r r }
    \hline
    prob. & cG-BLEMO & PSP-BLEMO & prob. & cG-BLEMO & PSP-BLEMO \\
    \hline
    DS1 & 2548.36 	 & 882.20   &   	DS3D & 3008.60 	 & 1393.23	\\
    DS2 & 2590.75 	 & 768.05 	&   	DS4 & 647.11 	 & 713.95\\
    DS3 & 2723.26 	 & 1278.38 	&	    DS5 & 560.06	 & 664.12\\
    DS1D & 2443.25 	 & 897.43 	&  	   TP1 & 2117.40 	 & 165.95 \\
    DS2D & 3159.10 	 & 785.70  &        TP2 & 1792.41 	 & 178.68\\
\hline
    \end{tabular}
    \label{tab:runtime}
\end{table}

SVM-BLEMO also uses LL PS prediction. However, its performance is significantly worse than PSP-BLEMO, as evident from Table~\ref{tab:compare_metric}. The IGD values are much higher than rest of the compared methods. This suggests that the LL search results of SVM-BLEMO are relatively further away from the optimal. Due to lower accuracy of LL PS prediction, it is not ideal for SVM based predictor to skip LL search entirely~(for up to $\gamma$ generations) in the proposed framework.

Four other state-of-the-art algorithms, namely BLMOCC, H-BLEMO, MOBEA-DPL and SMS-MOBO, are less similar conceptually to PSP-BLEMO. BLMOCC adopts co-evolution on both levels, while H-BLEMO involves hybridization via local search. One can observe from Table~\ref{tab:compare_metric} that PSP-BLEMO has lower mean IGD values than BLMOCC and H-BLEMO. In terms of FE, compared to BLMOCC, PSP-BLEMO uses fewer FE on the LL, and and more FE on the UL. However, the FE consumption on LL is usually 10 to 100 more times than UL (seen from Table~\ref{tab:compare_FE}). Thus, overall, PSP-BLEMO holds advantage over BLMOCC in both IGD metric and FE consumption. Similar observations are made with respect to H-BLEMO; additionally, PSP-BLEMO shows fewer evaluations also on the UL. MOBEA-DPL adopts two populations on the LL search, of which one population is evaluated on the UL. Therefore deceptive problems are challenging for MOBEA-DPL.  In terms of FE consumption, it uses up to 30 times more FE than PSP-BLEMO on the UL and up to 14 times more FE on the LL. SMS-MOBO shows competitive performance in 4 problems, but its performance needs up to 100 times more FEs~(on the LL) to achieve this performance. SMS-MOBO also faces challenges on deceptive problems, where it uses more FEs but returns rather high IGD values. Compared to SMS-MOBO, PSP-BLEMO shows stable performance across different problem types.

\begin{figure}[!h]
    \centering
    \subfloat[]{
    \includegraphics[width=0.4\textwidth]{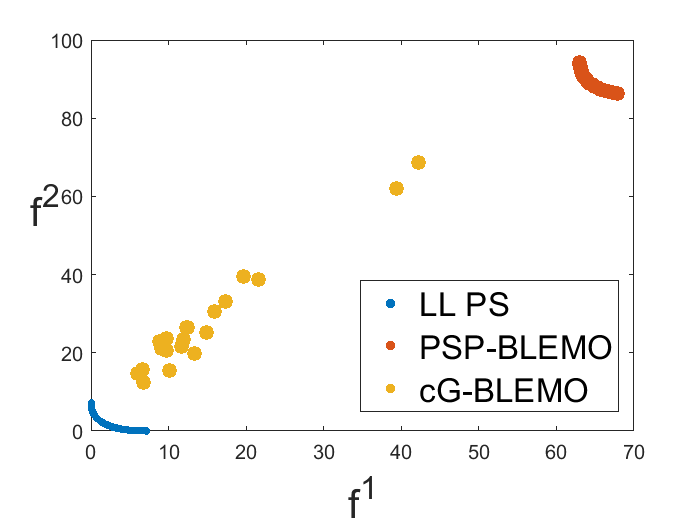}
    \label{fig:ds1gen1}}
    \subfloat[]{
    \includegraphics[width=0.4\textwidth]{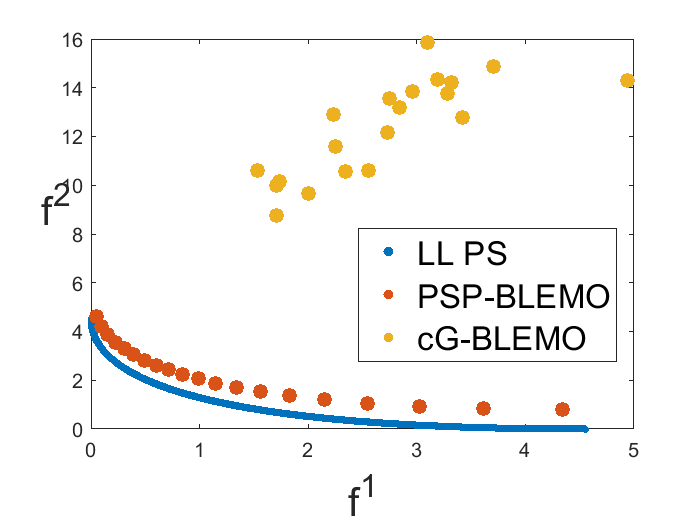}
    \label{fig:ds1gen3}}
 \caption{Comparison of LL PS prediction for a same~(randomly generated) UL solution between PSP-BLEMO and cG-BLEMO. (a) first generation, 400 training points for PSP and cG-BLEMO; (b) 1000 training points for PSP and 800 training points for cG-BLEMO}
 \label{fig:ds1LLcompare}
\end{figure}

Lastly, we explicitly observed the quality of the LL PS predictions, to account for the improvements in performance achieved by PSP-BLEMO. In Fig.~\ref{fig:ds1LLcompare}, we compared LL prediction results on PSP-BLEMO and cG-BLEMO. In Fig.~\ref{fig:ds1gen1}, it shows the prediction results obtained by two algorithms after first UL generation. PSP-BLEMO's prediction is not close to PF for this case. The reason is that after the first generation training data accumulated was only 400~(70\% training data therefore uses 280 data points). The size of training data is not enough to capture the UL solution to LL PS mapping. cG-BLEMO's PS prediction is relatively better in terms of being closer to PF. However, when the training data size increases to 1000~(70\% training is 700 data points), PSP-BLEMO's prediction is much closer to the true LL PF and well distributed as shown in Fig.~\ref{fig:ds1gen3}. As for cG-BLEMO, its prediction improves, but there is still significant amount of search effort required to move towards LL PF.
To summarize, PSP-BLEMO shows competitive performance in terms of UL IGD, FE consumption, and average runtime when compared with other state-of-the-art algorithms. Especially, when the target problem has deceptive nature and the number of allowable evaluations is relatively low. The competence of  PSP-BLEMO is supported by the LL PS prediction quality, which helps it skip  LL search for most of the generations with low impact on the quality of solutions obtained.

\subsection{Comparison study with `one-shot' PSP-BLEMO}

Above comparison with state-of-the-art algorithms shows competitive performance of PSP-BLEMO in terms of accuracy and FE consumption. The experimental settings used in previous studies~(e.g. population size 20) have been used for fair and consistent comparisons with the state-of-the-art algorithms. However, in this section, we would like to look into a variant of PSP-BLEMO and its performance when experimental settings and problem settings are varied. PSP-BLEMO, in its default form, updates NN every $\gamma$ generations. To attempt further savings in FE, one may consider a version of PSP-BLEMO that forgoes re-training entirely and train the model only once, in the first generation. We refer this version one-shot~(OS) PSP-BLEMO. During the run, the LL PS is always predicted, never obtained through search. At the end of the run, for the final reporting, the solutions of the last population are evaluated by running predicted LL assisted search on the ND solutions in UL archive. This process helps improve the solution quality on the LL, to more accurately reflect the quality of returned UL solutions. For a fair comparison, such a re-evaluation scheme is also applied to PSP-BLEMO in these experiments. 
In what follows, we use the shorthand OS and PSP to distinguish these two variants. Further, as a baseline, we use nested evolutionary algorithm~(NE), where both UL and LL search are carried out by MOEA described earlier in Section~\ref{sec:experiment} without the use of any predictions. For test problems, we expand the test bed to include problems of different variable sizes, so that the scalability of algorithms can also be observed. The detailed problem variable settings can be found in Table~\ref{tab:scalable_problems}.

\begin{table}[!ht]\footnotesize
    \centering
\caption{Three settings~(S1, S2 and S3) used for investigation. $D_u$ and $D_l$ refer to UL and LL variable size, respectively.}
\renewcommand\arraystretch{0.8}
  \begin{tabular}{l  l r r  r r }
    \hline
    ID & Prob. & Set 1 (S1) &  Set 2 (S2) & Set 3 (S3)\\
    \cline{3-5}
    && $D_u$/$D_l$ & $D_u$/$D_l$ & $D_u$/$D_l$ \\
    \hline
     1&   TP1 & 1/2 &  -   &  -\\
     2&   TP2 & 1/2 &  1/4 &  1/10\\
     3&   DS1 & 2/2 &  4/4 &  10/10\\
     4&   DS2 & 2/2 &  4/4 &  10/10\\
     5&   DS3  & 3/3 &  4/4 &  10/10\\
     6&   DS1D & 2/2 &  4/4 &  10/10\\
     7&   DS2D & 2/2 &  4/4 &  10/10\\
     8&   DS3D & 3/3 &  4/4 &  10/10\\
     9&  DS4 & 1/3  & 1/5 &   1/9 \\
     10 & DS5 & 1/3 & 1/5 &  1/9\\
    \hline
    \end{tabular}
    \label{tab:scalable_problems}
\end{table}

\begin{table*}[ht!]\scriptsize
\caption{Results using population size $20$ and IGD stopping at both levels. Statistical significance of the current algorithm w.r.t algorithm tagged $(i)$ is shown using symbols $\uparrow_{i}$, $\downarrow_{i}$ and $\approx_{i}$, indicating better, worse or equivalent performance, respectively}
    \centering
    \renewcommand\arraystretch{0.8}
    \begin{tabular}{l  r rr rrr |r rr rrr| rrr}
        \hline
        prob. &  NE(1) &  OS(2) & & PSP(3)  & & &   NE(1) & OS(2) & & PSP(3) & &  & NE & OS/NE & PSP/NE\\
        \hline
      \multicolumn{7}{c|}{UL Median IGD  (S1) } &  \multicolumn{6}{c}{LL normalized Median IGD  (S1) } &  \multicolumn{3}{|c}{  Median total FE  (S1) }\\
\hline
TP1 	& 0.0141 	& 0.0259 	& $\downarrow_1$ 	& \textbf{0.0128} 	& $\uparrow_1$ 	& $\uparrow_2$ 	& 0.8356 	& 0.4890 	& $\uparrow_1$ 	& \textbf{0.0746} 	& $\uparrow_1$ 	& $\uparrow_2$ 	& 387,414 	& 0.35 	& 0.66 	\\
TP2 	& 0.0251 	& 0.0193 	& $\approx_1$ 	& \textbf{0.0159} 	& $\uparrow_1$ 	& $\approx_2$ 	& 0.9551 	& \textbf{0.0799} 	& $\uparrow_1$ 	& 0.2247 	& $\uparrow_1$ 	& $\downarrow_2$ 	& 393,447 	& 0.52 	& 0.62 	\\
DS1 	& 0.0036 	& \textbf{0.0029} 	& $\uparrow_1$ 	& 0.0035 	& $\approx_1$ 	& $\downarrow_2$ 	& 0.9641 	& 0.1641 	& $\uparrow_1$ 	& \textbf{0.1041} 	& $\uparrow_1$ 	& $\approx_2$ 	& 446,479 	& 0.40 	& 0.58 	\\
DS2 	& 0.0262 	& \textbf{0.0241} 	& $\approx_1$ 	& 0.0252 	& $\approx_1$ 	& $\approx_2$ 	& \textbf{0.0004} 	& 0.0179 	& $\downarrow_1$ 	& 0.0103 	& $\downarrow_1$ 	& $\approx_2$ 	& 530,699 	& 0.32 	& 0.52 	\\
DS3 	& 0.0273 	& 0.0470 	& $\downarrow_1$ 	& \textbf{0.0198} 	& $\uparrow_1$ 	& $\uparrow_2$ 	& 0.8000 	& 0.6636 	& $\uparrow_1$ 	& \textbf{0.1143} 	& $\uparrow_1$ 	& $\uparrow_2$ 	& 817,075 	& 0.19 	& 0.33 	\\
DS1D 	& 0.0033 	& \textbf{0.0031}	& $\uparrow_1$ 	& \textbf{0.0031} 	& $\uparrow_1$ 	& $\approx_2$ 	& 0.0296 	& 0.0048 	& $\uparrow_1$ 	& \textbf{0.0025} 	& $\uparrow_1$ 	& $\uparrow_2$ 	& 472,344 	& 0.38 	& 0.57 	\\
DS2D 	& \textbf{0.0230} 	& 0.0248 	& $\approx_1$ 	& 0.0248 	& $\approx_1$ 	& $\approx_2$ 	& \textbf{0.0008} 	& 0.0274 	& $\approx_1$ 	& 0.0014 	& $\approx_1$ 	& $\uparrow_2$ 	& 555,323 	& 0.31 	& 0.49 	\\
DS3D 	& 0.0313 	& 0.0389 	& $\approx_1$ 	& \textbf{0.0241} 	& $\approx_1$ 	& $\uparrow_2$ 	& 0.6865 	& 0.5016 	& $\uparrow_1$ 	& \textbf{0.0797} 	& $\uparrow_1$ 	& $\uparrow_2$ 	& 835,690 	& 0.19 	& 0.31 	\\
DS4 	& 0.0365 	& 0.0206 	& $\uparrow_1$ 	& \textbf{0.0102} 	& $\uparrow_1$ 	& $\uparrow_2$ 	& 0.8708 	& 0.0176 	& $\uparrow_1$ 	& \textbf{0.0163} 	& $\uparrow_1$ 	& $\approx_2$ 	& 244,766 	& 1.69 	& 7.06 	\\
DS5 	& 0.0336 	& 0.0178 	& $\uparrow_1$ 	& \textbf{0.0086} 	& $\uparrow_1$ 	& $\uparrow_2$ 	& 0.8210 	& \textbf{0.0129} 	& $\uparrow_1$ 	& 0.0232 	& $\uparrow_1$ 	& $\downarrow_2$ 	& 254,998 	& 1.47 	& 5.86 	\\
\hline
      \multicolumn{7}{c|}{UL Median IGD  (S2) } &  \multicolumn{6}{c}{LL normalized Median IGD  (S2) } &  \multicolumn{3}{|c}{  Median total FE  (S2) }\\
\hline
TP2 	& 0.0393 	& 0.0272 	& $\uparrow_1$ 	& \textbf{0.0192} 	& $\uparrow_1$ 	& $\uparrow_2$ 	& 0.9351 	& \textbf{0.0488} 	& $\uparrow_1$ 	& 0.1842 	& $\uparrow_1$ 	& $\downarrow_2$ 	& 559,646 	& 0.36 	& 0.42 	\\
DS1 	& 0.0077 	& 0.0062 	& $\uparrow_1$ 	& \textbf{0.0052} 	& $\uparrow_1$ 	& $\approx_2$ 	& 0.9322 	& 0.6535 	& $\uparrow_1$ 	& \textbf{0.0980} 	& $\uparrow_1$ 	& $\uparrow_2$ 	& 1,407,780 	& 0.14 	& 0.24 	\\
DS2 	& 0.0391 	& 0.0425 	& $\approx_1$ 	& \textbf{0.0336} 	& $\approx_1$ 	& $\uparrow_2$ 	& \textbf{0.0001} 	& \textbf{0.0001} 	& $\approx_1$ 	& 0.0003 	& $\approx_1$ 	& $\approx_2$ 	& 1,554,432 	& 0.12 	& 0.24 	\\
DS3 	& \textbf{0.0424} 	& 0.0524 	& $\approx_1$ 	& 0.0468 	& $\approx_1$ 	& $\approx_2$ 	& 0.1730 	& 0.2363 	& $\downarrow_1$ 	& \textbf{0.0445} 	& $\approx_1$ 	& $\uparrow_2$ 	& 1,383,181 	& 0.12 	& 0.22 	\\
DS1D 	& 0.0068 	& 0.0054 	& $\approx_1$ 	& \textbf{0.0046} 	& $\uparrow_1$ 	& $\approx_2$ 	& 0.8856 	& 0.5674 	& $\uparrow_1$ 	& \textbf{0.0714} 	& $\uparrow_1$ 	& $\uparrow_2$ 	& 1,281,886 	& 0.16 	& 0.27 	\\
DS2D 	& \textbf{0.0330} 	& 0.0351 	& $\approx_1$ 	& 0.0393 	& $\approx_1$ 	& $\approx_2$ 	& 0.0000 	& 0.0000 	& $\approx_1$ 	& \textbf{0.0000} 	& $\approx_1$ 	& $\uparrow_2$ 	& 1,622,311 	& 0.11 	& 0.23 	\\
DS3D 	& \textbf{0.0429} 	& 0.0453 	& $\approx_1$ 	& 0.0550 	& $\approx_1$ 	& $\approx_2$ 	& 0.3411 	& 0.3884 	& $\downarrow_1$ 	& \textbf{0.0974} 	& $\uparrow_1$ 	& $\uparrow_2$ 	& 1,375,028 	& 0.12 	& 0.21 	\\
DS4 	& 0.1248 	& 0.0304 	& $\uparrow_1$ 	& \textbf{0.0102} 	& $\uparrow_1$ 	& $\uparrow_2$ 	& 0.8762 	& 0.0383 	& $\uparrow_1$ 	& \textbf{0.0150} 	& $\uparrow_1$ 	& $\uparrow_2$ 	& 393,252 	& 0.81 	& 4.21 	\\
DS5 	& 0.1128 	& 0.0280 	& $\uparrow_1$ 	& \textbf{0.0088} 	& $\uparrow_1$ 	& $\uparrow_2$ 	& 0.8920 	& 0.0392 	& $\uparrow_1$ 	& \textbf{0.0167} 	& $\uparrow_1$ 	& $\uparrow_2$ 	& 454,015 	& 0.63 	& 3.00 	\\
\hline
      \multicolumn{7}{c|}{UL Median IGD  (S3) } &  \multicolumn{6}{c}{LL normalized Median IGD  (S3) } &  \multicolumn{3}{|c}{  Median total FE  (S3) }\\
\hline
TP2 	& 0.0475 	& 0.0447 	& $\approx_1$ 	& \textbf{0.0229} 	& $\uparrow_1$ 	& $\uparrow_2$ 	& 0.9464 	& \textbf{0.0628} 	& $\uparrow_1$ 	& 0.0975 	& $\uparrow_1$ 	& $\downarrow_2$ 	& 992,290 	& 0.20 	& 0.24 	\\
DS1 	& 0.0296 	& 9.5274 	& $\downarrow_1$ 	& \textbf{0.0177} 	& $\uparrow_1$ 	& $\uparrow_2$ 	& 0.4790 	& 0.5024 	& $\approx_1$ 	& \textbf{0.0736} 	& $\uparrow_1$ 	& $\uparrow_2$ 	& 7,796,493 	& 0.04 	& 0.08 	\\
DS2 	& 0.0962 	& 7.5965 	& $\downarrow_1$ 	& \textbf{0.0938} 	& $\approx_1$ 	& $\uparrow_2$ 	& 0.0206 	& \textbf{0.0000} 	& $\uparrow_1$ 	& 0.0615 	& $\approx_1$ 	& $\downarrow_2$ 	& 8,405,941 	& 0.04 	& 0.08 	\\
DS3 	& 0.1186 	& 38.2745 	& $\downarrow_1$ 	& \textbf{0.1172}	& $\approx_1$ 	& $\uparrow_2$ 	& 0.7870 	& \textbf{0.2159} 	& $\uparrow_1$ 	& 0.4140 	& $\uparrow_1$ 	& $\downarrow_2$ 	& 9,183,774 	& 0.01 	& 0.07 	\\
DS1D 	& 0.0445 	& 873.5308 	& $\downarrow_1$ 	& \textbf{0.0177} 	& $\uparrow_1$ 	& $\uparrow_2$ 	& 0.0295 	& 0.0115 	& $\uparrow_1$ 	& \textbf{0.0026} 	& $\uparrow_1$ 	& $\uparrow_2$ 	& 6,428,462 	& 0.10 	& 0.09 	\\
DS2D 	& 0.1157 	& 780.8078 	& $\downarrow_1$ 	& \textbf{0.0877} 	& $\uparrow_1$ 	& $\uparrow_2$ 	& 0.0025 	& \textbf{0.0000} 	& $\uparrow_1$ 	& 0.0117	& $\approx_1$ 	& $\downarrow_2$ 	& 6,454,517 	& 0.07 	& 0.10 	\\
DS3D 	& \textbf{0.1125} 	& 112.9853 	& $\downarrow_1$ 	& 0.1572	& $\approx_1$ 	& $\uparrow_2$ 	& 0.8029 	& 0.5159 	& $\uparrow_1$ 	& \textbf{0.3692} 	& $\uparrow_1$ 	& $\uparrow_2$ 	& 8,660,013 	& 0.05 	& 0.07 	\\
DS4 	& 1.0114 	& 0.0339 	& $\uparrow_1$ 	& \textbf{0.0119} 	& $\uparrow_1$ 	& $\uparrow_2$ 	& 0.6530 	& 0.1837 	& $\uparrow_1$ 	& \textbf{0.0273} 	& $\uparrow_1$ 	& $\uparrow_2$ 	& 693,326 	& 0.39 	& 2.34 	\\
DS5 	& 1.0475 	& 0.0341 	& $\uparrow_1$ 	& \textbf{0.0098} 	& $\uparrow_1$ 	& $\uparrow_2$ 	& 0.8778 	& 0.2817 	& $\uparrow_1$ 	& \textbf{0.0460} 	& $\uparrow_1$ 	& $\uparrow_2$ 	& 799,228 	& 0.30 	& 1.77 	\\
\hline
        \end{tabular}
        \label{tab:batch3_20popsetting_pspeval}
\end{table*}

\begin{table*}[ht!]\scriptsize
\caption{Results using population size and maximum generation size $10D$, and IGD stopping condition on the both levels}
    \centering
    \renewcommand\arraystretch{0.8}
    \begin{tabular}{l  r rr rrr |r rr rrr| rrr}
        \hline
        prob. &  NE(1) &  OS(2) & & PSP(3)  & & &   NE(1) & OS(2) & & PSP(3) & &  & NE & OS/NE & PSP/NE\\
        \hline
\multicolumn{7}{c|}{UL Median IGD  (S1) } &  \multicolumn{6}{c}{LL normalized Median IGD  (S1) } & \multicolumn{3}{|c}{ Median total FE  (S1) }\\
\hline        
TP1 	& 0.0229 	& 0.0423 	& $\downarrow_1$ 	& \textbf{0.0182} 	& $\uparrow_1$ 	& $\uparrow_2$ 	& 0.8104 	& 0.4963 	& $\uparrow_1$ 	& \textbf{0.0449} 	& $\uparrow_1$ 	& $\uparrow_2$ 	& 40,268 	& 0.33 	& 1.16 	\\
TP2 	& 0.0357 	& 0.0290 	& $\uparrow_1$ 	& \textbf{0.0265} 	& $\uparrow_1$ 	& $\approx_2$ 	& 0.9032 	& \textbf{0.0934}	& $\uparrow_1$ 	& 0.2158 	& $\uparrow_1$ 	& $\downarrow_2$ 	& 41,332 	& 0.62 	& 1.14 	\\
DS1 	& 0.0052 	& \textbf{0.0043} 	& $\uparrow_1$ 	& 0.0049 	& $\approx_1$ 	& $\downarrow_2$ 	& 0.7498 	& 0.1824 	& $\uparrow_1$ 	& \textbf{0.0525} 	& $\uparrow_1$ 	& $\uparrow_2$ 	& 163,905 	& 0.29 	& 0.76 	\\
DS2 	& 0.0416 	& 0.0392 	& $\approx_1$ 	& \textbf{0.0353} 	& $\uparrow_1$ 	& $\approx_2$ 	& 0.3350 	& 0.3142 	& $\approx_1$ 	& \textbf{0.0106} 	& $\uparrow_1$ 	& $\uparrow_2$ 	& 162,795 	& 0.17 	& 0.76 	\\
DS3 	& 0.0314 	& 0.0241 	& $\approx_1$ 	& \textbf{0.0208} 	& $\uparrow_1$ 	& $\approx_2$ 	& 0.8655 	& 0.2220 	& $\uparrow_1$ 	& \textbf{0.0718} 	& $\uparrow_1$ 	& $\uparrow_2$ 	& 802,375 	& 0.12 	& 0.29 	\\
DS1D 	& 0.0055 	& \textbf{0.0044} 	& $\uparrow_1$ 	& 0.0047 	& $\uparrow_1$ 	& $\approx_2$ 	& 0.0337 	& 0.0082 	& $\uparrow_1$ 	& \textbf{0.0025} 	& $\uparrow_1$ 	& $\uparrow_2$ 	& 163,735 	& 0.28 	& 0.75 	\\
DS2D 	& 0.0396 	& 0.0384 	& $\approx_1$ 	& \textbf{0.0328} 	& $\uparrow_1$ 	& $\uparrow_2$ 	& 0.2660 	& 0.1616 	& $\uparrow_1$ 	& \textbf{0.0070} 	& $\uparrow_1$ 	& $\uparrow_2$ 	& 162,068 	& 0.17 	& 0.77 	\\
DS3D 	& 0.0281 	& 0.0261 	& $\approx_1$ 	& \textbf{0.0215} 	& $\uparrow_1$ 	& $\approx_2$ 	& 0.4273 	& 0.1088 	& $\uparrow_1$ 	& \textbf{0.0265} 	& $\uparrow_1$ 	& $\uparrow_2$ 	& 753,722 	& 0.13 	& 0.33 	\\
DS4 	& 0.0574 	& 0.0189 	& $\uparrow_1$ 	& \textbf{0.0115} 	& $\uparrow_1$ 	& $\uparrow_2$ 	& 0.8489 	& 0.0518 	& $\uparrow_1$ 	& \textbf{0.0478} 	& $\uparrow_1$ 	& $\approx_2$ 	& 53,404 	& 6.24 	& 20.37 	\\
DS5 	& 0.0523 	& 0.0136 	& $\uparrow_1$ 	& \textbf{0.0108} 	& $\uparrow_1$ 	& $\uparrow_2$ 	& 0.8562 	& 0.0452 	& $\uparrow_1$ 	& \textbf{0.0442} 	& $\uparrow_1$ 	& $\approx_2$ 	& 52,822 	& 3.74 	& 15.63 \\
\hline
\multicolumn{7}{c|}{UL Median IGD  (S2) } &  \multicolumn{6}{c}{LL normalized Median IGD  (S2) } & \multicolumn{3}{|c}{ Median total FE  (S2) }\\
\hline
TP2 	& 0.0391 	& 0.0313 	& $\uparrow_1$ 	& \textbf{0.0255} 	& $\uparrow_1$ 	& $\uparrow_2$ 	& 0.9437 	& \textbf{0.0699} 	& $\uparrow_1$ 	& 0.2299 	& $\uparrow_1$ 	& $\downarrow_2$ 	& 156,299 	& 0.37 	& 0.65 	\\
DS1 	& 0.0053 	& \textbf{0.0044}	& $\uparrow_1$ 	& \textbf{0.0044} 	& $\uparrow_1$ 	& $\approx_2$ 	& 0.9173 	& 0.1746 	& $\uparrow_1$ 	& \textbf{0.0424} 	& $\uparrow_1$ 	& $\uparrow_2$ 	& 2,573,072 	& 0.10 	& 0.21 	\\
DS2 	& 0.0483 	& 0.0488 	& $\approx_1$ 	& \textbf{0.0441} 	& $\uparrow_1$ 	& $\approx_2$ 	& 0.2119 	& 0.0871 	& $\uparrow_1$ 	& \textbf{0.0101}	& $\uparrow_1$ 	& $\uparrow_2$ 	& 2,532,996 	& 0.08 	& 0.16 	\\
DS3 	& 0.0416 	& \textbf{0.0302} 	& $\uparrow_1$ 	& 0.0348 	& $\approx_1$ 	& $\approx_2$ 	& 0.8972 	& 0.2186 	& $\uparrow_1$ 	& \textbf{0.0445} 	& $\uparrow_1$ 	& $\uparrow_2$ 	& 2,419,620 	& 0.08 	& 0.18 	\\
DS1D 	& 0.0047 	& \textbf{0.0038} 	& $\uparrow_1$ 	& 0.0053 	& $\approx_1$ 	& $\downarrow_2$ 	& 0.9688 	& 0.1734 	& $\uparrow_1$ 	& \textbf{0.0516}	& $\uparrow_1$ 	& $\uparrow_2$ 	& 2,570,759 	& 0.10 	& 0.21 	\\
DS2D 	& \textbf{0.0400} 	& 0.0415 	& $\approx_1$ 	& 0.0454 	& $\approx_1$ 	& $\approx_2$ 	& 0.1685 	& 0.0709 	& $\uparrow_1$ 	& \textbf{0.0066} 	& $\uparrow_1$ 	& $\uparrow_2$ 	& 2,529,416 	& 0.07 	& 0.16 	\\
DS3D 	& 0.0276 	& 0.0360 	& $\downarrow_1$ 	& \textbf{0.0257} 	& $\approx_1$ 	& $\uparrow_2$ 	& 0.9083 	& 0.2218 	& $\uparrow_1$ 	& \textbf{0.0304} 	& $\uparrow_1$ 	& $\uparrow_2$ 	& 2,527,698 	& 0.08 	& 0.19 	\\
DS4 	& 0.1663 	& 0.0282 	& $\uparrow_1$ 	& \textbf{0.0083} 	& $\uparrow_1$ 	& $\uparrow_2$ 	& 0.8702 	& 0.1118 	& $\uparrow_1$ 	& \textbf{0.0216} 	& $\uparrow_1$ 	& $\uparrow_2$ 	& 124,884 	& 2.73 	& 14.51 	\\
DS5 	& 0.1799 	& 0.0201 	& $\uparrow_1$ 	& \textbf{0.0075} 	& $\uparrow_1$ 	& $\uparrow_2$ 	& 0.8563 	& 0.1138 	& $\uparrow_1$ 	& \textbf{0.0309} 	& $\uparrow_1$ 	& $\uparrow_2$ 	& 123,403 	& 2.58 	& 10.78 	\\
\hline
\multicolumn{7}{c|}{UL Median IGD  (S3) } &  \multicolumn{6}{c}{LL normalized Median IGD  (S3) } & \multicolumn{3}{|c}{ Median total FE  (S3) }\\
\hline
TP2 	& 0.0388 	& 0.0310 	& $\uparrow_1$ 	& \textbf{0.0251} 	& $\uparrow_1$ 	& $\uparrow_2$ 	& 0.9035 	& \textbf{0.0938} 	& $\uparrow_1$ 	& 0.1116 	& $\uparrow_1$ 	& $\approx_2$ 	& 926,877 	& 0.22 	& 0.29 	\\
DS1 	& 0.0622 	& 0.0403 	& $\uparrow_1$ 	& \textbf{0.0376} 	& $\uparrow_1$ 	& $\approx_2$ 	& 0.9457 	& 0.1131 	& $\uparrow_1$ 	& \textbf{0.0084} 	& $\uparrow_1$ 	& $\uparrow_2$ 	& 99,386,090 	& 0.03 	& 0.07 	\\
DS2 	& 0.1612 	& 0.1260 	& $\uparrow_1$ 	& \textbf{0.1098} 	& $\uparrow_1$ 	& $\uparrow_2$ 	& 0.0247 	& 0.0092 	& $\uparrow_1$ 	& \textbf{0.0000} 	& $\uparrow_1$ 	& $\uparrow_2$ 	& 98,265,642 	& 0.03 	& 0.06 	\\
DS3 	& 0.1536 	& \textbf{0.1007} 	& $\uparrow_1$ 	& 0.1540 	& $\approx_1$ 	& $\downarrow_2$ 	& 0.9055 	& \textbf{0.0722} 	& $\uparrow_1$ 	& 0.0733 	& $\uparrow_1$ 	& $\approx_2$ 	& 93,243,306 	& 0.03 	& 0.07 	\\
DS1D 	& 2.5416 	& \textbf{0.0358} 	& $\uparrow_1$ 	& 0.0505 	& $\approx_1$ 	& $\downarrow_2$ 	& 0.2302 	& 0.0017 	& $\uparrow_1$ 	& \textbf{0.0002} 	& $\uparrow_1$ 	& $\uparrow_2$ 	& 99,369,828 	& 0.03 	& 0.07 	\\
DS2D 	& 2.0997 	& 0.1153 	& $\uparrow_1$ 	& \textbf{0.1054} 	& $\uparrow_1$ 	& $\approx_2$ 	& 0.0000 	& 0.0005 	& $\approx_1$ 	& \textbf{0.0000} 	& $\uparrow_1$ 	& $\uparrow_2$ 	& 98,163,613 	& 0.03 	& 0.06 	\\
DS3D 	& 0.1853 	& \textbf{0.0768} 	& $\uparrow_1$ 	& 0.1544 	& $\approx_1$ 	& $\downarrow_2$ 	& 0.0246 	& \textbf{0.0006} 	& $\uparrow_1$ 	& \textbf{0.0006} 	& $\uparrow_1$ 	& $\approx_2$ 	& 87,311,054 	& 0.03 	& 0.08 	\\
DS4 	& 1.4533 	& 0.0334 	& $\uparrow_1$ 	& \textbf{0.0090} 	& $\uparrow_1$ 	& $\uparrow_2$ 	& 0.9303 	& 0.5862 	& $\uparrow_1$ 	& \textbf{0.1425} 	& $\uparrow_1$ 	& $\uparrow_2$ 	& 394,536 	& 1.15 	& 6.37 	\\
DS5 	& 1.4162 	& 0.0435 	& $\uparrow_1$ 	& \textbf{0.0088} 	& $\uparrow_1$ 	& $\uparrow_2$ 	& 0.8635 	& 0.3760 	& $\uparrow_1$ 	& \textbf{0.2459} 	& $\uparrow_1$ 	& $\uparrow_2$ 	& 393,729 	& 0.95 	& 4.52 	\\
\hline
        \end{tabular}
        \label{tab:batch3_10dsetting_pspeval}
\end{table*}

In the first part of this experiment, we keep the population size fixed to 20, as done previously. The performance in terms of normalized UL IGD, normalized LL IGD and total FE are summarized in Table~\ref{tab:batch3_20popsetting_pspeval}. The calculation of normalized LL IGD is as follows. Firstly, for every UL search result, we compute the normalized LL IGD for each UL solution. Next, the mean of these normalized LL IGD values is associated with each search result. Subsequently, we identify the maximum and minimum normalized IGD values across all three algorithms~(considering all seeds) to establish the range for normalizing each LL IGD result between 0 and 1. The reported normalized LL IGD is the final result from above 3 steps.
Statistical test results are shown in symbols $\uparrow_{i}$, $\downarrow_{i}$ and $\approx_{i}$. This $i$ means compared to $(i)$ algorithm, current algorithm~(on the left of symbols) performs better, worse or equivalent. For FE consumption, except for DS4 and DS5, it is not surprising that OS is the best, and PSP uses more FE than OS, but both of them use fewer FE than the baseline.
The larger the number of variables, the more savings on FE the proposed method can achieve. 
For DS4 and DS5, due to additional search on the UL, PSP has more FE consumed than baseline, but this number decreases as the number of variables increases as shown in the last section of Table~\ref{tab:batch3_20popsetting_pspeval}. For OS, its FE is higher than baseline when variable size is small~(S1), but in the rest of the settings, its FE consumption becomes much smaller than baseline and PSP.

It can be noted that PSP shows better performance in all variable settings. For S1, PSP performs equal to or better than baseline in all problems on the UL, 9 out of 10 on the LL. For S2 and S3, PSP performs equal or better than baseline~ in all problems on UL and LL. Between PSP and OS, similar trend can be observed on the UL. 
For all problems except one across three settings, PSP performs better than or equal to OS. 
LL situation varies mainly in S3 section, where where OS outperforms PSP 4 out of 10 cases on LL. But when we check the corresponding UL performance, e.g. DS2, DS3 and DS2D etc, OS has much worse UL IGD values. Overall, PSP presents most consistent performance in this setting. The performance of OS is also inline with expectation, i.e. with a population size of 20 and for small number of variables, it would have accumulated enough training data for appropriate prediction accuracy. However, for larger number of variables, its prediction accuracy is compromised affecting the overall performance.

In the second part of this experiment, we set the population size to be $10D$ and maximum generation size to be $10D$~(while also retaining the IGD termination criterion), making them proportional to the problem variable size. $D$ refers to corresponding level variable size. For UL, it means $D_u$, while for LL it means $D_l$. Table~\ref{tab:batch3_10dsetting_pspeval} shows results of these experiments. Again, it is seen that OS and PSP use much fewer FE compared to the baseline except DS4 and DS5. For DS4 and DS5, similar to the first experiment, when problem variable size increases, this extra FE consumption also decreases compared to those used by baseline. Between PSP and baseline, on both levels and across all problem settings, PSP performs significantly better.  
Between PSP and OS, the performance is not as overwhelmingly in favor of PSP as it has been in the previous experiments.

For S1 and S2, PSP still shows better performance on both levels, while for S3, PSP performs better than OS at UL in 4 problems and worse in 3 problems. On the LL, PSP performs better or equal to OS in all cases. For S3, the population size and generations are significantly larger in this case~(e.g. $100$ for $10$ variables). Therefore, OS has same number of training points as PSP, except that PSP can also update the models later. The competitive performance of OS suggests that the proposed LL PS prediction model, being built on a larger dataset, can maintain its performance quite well over generations. This observation actually also resonates with the Table~\ref{tab:keyparameter_sigtest} last column about test results on $\gamma$ when $ds=5e3$. Although according to wining cases, we choose $\gamma=10$ as our parameter value, for majority of the cases, the differences are not significant~(i.e. majority of the cases are `equivalent'), $\gamma=10$ is only marginally better than others. So, the OS performance can be competitive when $ds$ is large. Overall, we still tend to choose PSP with re-training~(every $\gamma$ generations) as the default. Although the FE consumption on DS4 and DS5 is high, it is still competitive with the state-of-the-art, and such problems are not too common in BLMOPs. Considering the performance at both UL and LL, it is the most consistent and competitive strategy across different settings and provides an opportunity for model correction intermittently. But the insights gained from this last set of experiments provides an avenue for further FE reduction. The use of OS strategy can be beneficial towards this endeavor without significant compromise on performance, provided a sufficiently large dataset is used to train the models.

\section{Conclusion and future work}
\label{sec:conclusion}
Solving BLMOPs using a nested approach requires exorbitant number of function evaluations. Therefore, it is of interest to develop techniques that can reduce the computational expense while achieving good solution quality. In this paper, we develop a Pareto set prediction assisted search~(PSP-BLEMO) towards addressing this challenge. To achieve this, the archive of evaluated solutions is transformed using a helper variable, and an NN model is trained on the resulting dataset. The model is then used to predict the PS approximation, used to seed the LL population or bypass the LL search entirely. The model itself can be re-trained periodically using additional UL solutions that undergo LL search every $\gamma$ generations. For saving additional FEs, a `one-shot' approach can be applied where the model is not updated during the run.

Extensive numerical experiments were conducted to test the efficacy of the proposed approach in terms of, solution quality, FE consumption and runtime. Key parameters and variants of the algorithm were also analyzed. Comparisons with the state-of-the-art algorithms shows competitive and consistent performance of PSP-BLEMO. The performance is particularly notable where deceptive functions are involved. Moreover, it was concluded that the performance of the algorithm was not too sensitive to the training gap $\gamma$, as long as there is sufficient training data available to build accurate PSP models.

In the future work, further improvements in the algorithmic performance will be investigated, e.g., by enhancing the modeling through use of more carefully selected subset from the data. Currently, the most recent LL search results are used as training data. Its distribution might not be uniformly distributed over the entire PF. Decomposition based segmentation on objective space can provide opportunity in improving this data quality. 
Moreover, extensions for higher number of objectives could also be considered. The way in which helper variable $r$ is used in this study is particularly suited for two-objective problems, since arranging $r$ and $f_1$ corresponds to sequentially traversing the span of the PF. However, for problems with more objectives, the composition of $r$ will need to be further refined. One potential way is to utilize reference vectors obtained through systematically sampled points using normalized boundary intersection method, commonly utilized in decomposition-based MOEAs. In the current state of the field, more than two objectives are rare, hence this would be an interesting extension of this study in the future. Lastly, extension of some other recent modeling structures, such as hypernetworks, to solve BLMOPs could also be explored.


\bibliographystyle{ACM-Reference-Format}
\bibliography{main}

\end{document}


\title{--Supplementary material-- \\ Pareto Set Prediction Assisted Bilevel Multi-objective Optimization}




\makeatletter
\let\@authorsaddresses\@empty
\makeatother

\ccsdesc[500]{Computing methodologies~Continuous space search}
\keywords{Evolutionary optimization, bilevel optimization, Pareto set generation, neural networks}


\maketitle

\section{Experimental results for parameter analysis}

\begin{table*}[!ht]\footnotesize
\caption{}
    \centering
    \renewcommand\arraystretch{0.8}
    \setlength\tabcolsep{0.4mm} 
    \begin{tabular}{|l| r r r  |r r r |r r r | r r r| r r r|}
    \hline
        \multirow{2}{*}{Prob.} &\multicolumn{3}{c|}{$\gamma =5$} & \multicolumn{3}{c|}{$\gamma =10$ } &\multicolumn{3}{c|}{$\gamma =15$} & \multicolumn{3}{c|}{$\gamma =20$} & \multicolumn{3}{c|}{$\gamma =Inf$}  \\
        \cline{2-16}
        &  IGD & UL FE & LL FE &  IGD & $\frac{\text{UL FE}}{\gamma = 5}$ & $\frac{\text{LL FE}}{\gamma = 5}$ &  IGD & $\frac{\text{UL FE}}{\gamma = 5}$ &  $\frac{\text{LL FE}}{\gamma = 5}$ & IGD & $\frac{\text{UL FE}}{\gamma = 5}$ & $\frac{\text{LL FE}}{\gamma = 5}$ & IGD & $\frac{\text{UL FE}}{\gamma = 5}$ & $\frac{\text{UL FE}}{\gamma = 5}$\\
        \cline{1-16}
         \multicolumn{16}{|c|}{ Data size $ds=500$} \\
        \cline{1-16}
DS1 & 0.0237 	& 45,951 	 & 885,905 	 & 0.0214 	& 1.10 	 & 0.67 	 & \textbf{0.0186} 	& 1.07 	 & 0.54 	 & 0.0285 	& 1.00 	 & 0.47 	 & 13.5974 	& 1.22 	 & 0.28 	 \\
DS2 & \textbf{0.0905} 	& 42,627 	 & 750,250 	 & 0.0980 	& 0.96 	 & 0.69 	 & 0.1097 	& 0.95 	 & 0.59 	 & 0.1023 	& 0.98 	 & 0.52 	 & 10.4672 	& 1.01 	 & 0.29 	 \\
DS3 & \textbf{0.8498} 	& 28,148 	 & 3,027,181 	 & 16.8926 	& 0.09 	 & 0.13 	 & 26.9720 	& 0.06 	 & 0.09 	 & 35.5517 	& 0.04 	 & 0.06 	 & 34.6524 	& 0.04 	 & 0.06 	 \\
DS1D & 236.3090 	& 198,710 	 & 4,781,983 	 & 253.5029 	& 1.16 	 & 0.55 	 & 229.8114 	& 1.35 	 & 0.40 	 & \textbf{213.3356} 	& 1.40 	 & 0.29 	 & 588.5591 	& 0.40 	 & 0.10 	 \\
DS2D & \textbf{0.1192} 	& 45,303 	 & 748,687 	 & 7.3412 	& 2.94 	 & 1.93 	 & 37.6463 	& 5.80 	 & 1.57 	 & 32.2145 	& 0.63 	 & 0.44 	 & 386.4032 	& 1.12 	 & 0.42 	 \\
DS3D & 88.1326 	& 159,576 	 & 9,355,336 	 & \textbf{55.1305} 	& 0.09 	 & 0.12 	 & 115.9796 	& 0.38 	 & 0.38 	 & 97.3870 	& 0.10 	 & 0.11 	 & 164.2486 	& 0.07 	 & 0.04 	 \\
DS4 &  \textbf{0.0190} 	& 361895 	 & 191704 	 & 0.0233 	& 0.63 	 & 0.86 	 & 0.0240 	& 0.61 	 & 0.84 	 & 0.0282 	& 0.43 	 & 0.75 	 & 0.0282 	& 0.43 	 & 0.75 	 \\
DS5 &  \textbf{0.0184} 	& 345427 	 & 192346 	 & 0.0215 	& 0.61 	 & 0.86 	 & 0.0223 	& 0.57 	 & 0.82 	 & 0.0260 	& 0.39 	 & 0.74 	 & 0.0260 	& 0.39 	 & 0.74 	 \\
TP1 & \textbf{0.0154} 	& 6,976 	 & 166,241 	 & 0.0202 	& 0.99 	 & 0.89 	 & 0.0248 	& 0.79 	 & 0.87 	 & 0.0273 	& 0.77 	 & 0.82 	 & 0.0273 	& 0.77 	 & 0.82 	 \\
TP2 & \textbf{0.0283} 	& 9,878 	 & 167,175 	 & 0.0292 	& 1.19 	 & 0.93 	 & 0.0322 	& 1.19 	 & 0.89 	 & 0.0326 	& 0.72 	 & 0.81 	 & 0.0333 	& 0.72 	 & 0.79 	 \\
\cline{1-16}
         \multicolumn{16}{|c|}{ $ds = 1000$} \\
        \cline{1-16}
DS1 & 0.0195 	& 51,308 	 & 530,028 	 & \textbf{0.0172} 	& 1.10 	 & 0.82 	 & 0.0208 	& 1.00 	 & 0.70 	 & 0.0219 	& 0.95 	 & 0.66 	 & 0.0633 	& 0.88 	 & 0.50 	 \\
DS2 & 0.1173 	& 40,851 	 & 684,127 	 & 0.1128 	& 0.95 	 & 0.76 	 & \textbf{0.0997} 	& 1.01 	 & 0.63 	 & 0.1079 	& 0.99 	 & 0.54 	 & 0.3458 	& 0.85 	 & 0.37 	 \\
DS3 & 0.2035 	& 38,585 	 & 747,622 	 & 0.1609 	& 1.01 	 & 0.87 	 & \textbf{0.1585} 	& 1.08 	 & 0.73 	 & 0.2398 	& 1.12 	 & 0.73 	 & 2.2858 	& 0.78 	 & 0.41 	 \\
DS1D & 0.0194 	& 48,379 	 & 516,653 	 & \textbf{0.0141} 	& 1.11 	 & 0.83 	 & 0.0159 	& 1.11 	 & 0.74 	 & 0.0187 	& 1.01 	 & 0.66 	 & 0.0267 	& 0.95 	 & 0.52 	 \\
DS2D & 0.1086 	& 41,010 	 & 675,599 	 & \textbf{0.0930} 	& 0.99 	 & 0.73 	 & 0.1013 	& 0.92 	 & 0.60 	 & 0.0942 	& 0.94 	 & 0.56 	 & 0.1020 	& 0.83 	 & 0.37 	 \\
DS3D & \textbf{0.7827} 	& 32,905 	 & 741,202 	 & 14.5949 	& 2.74 	 & 1.32 	 & 9.4464 	& 7.65 	 & 1.65 	 & 12.1191 	& 0.94 	 & 0.67 	 & 4.5521 	& 0.68 	 & 0.44 	 \\
DS4 &  \textbf{0.0173} 	& 436391 	 & 198501 	 & 0.0207 	& 0.68 	 & 0.88 	 & 0.0208 	& 0.69 	 & 0.86 	 & 0.0230 	& 0.52 	 & 0.80 	 & 0.0230 	& 0.52 	 & 0.80 	 \\
DS5 &  \textbf{0.0177} 	& 410849 	 & 196273 	 & 0.0189 	& 0.67 	 & 0.88 	 & 0.0192 	& 0.66 	 & 0.88 	 & 0.0225 	& 0.50 	 & 0.80 	 & 0.0225 	& 0.50 	 & 0.80 	 \\
TP1 & \textbf{0.0161} 	& 7,337 	 & 173,606 	 & 0.0189 	& 0.78 	 & 0.88 	 & 0.0211 	& 0.79 	 & 0.87 	 & 0.0242 	& 0.75 	 & 0.82 	 & 0.0242 	& 0.75 	 & 0.82 	 \\
TP2 & 0.0297	& 7,891 	 & 165,926 	 & 0.0305 	& 1.00 	 & 0.91 	 & \textbf{0.0273} 	& 1.24 	 & 0.89 	 & 0.0290 	& 0.80 	 & 0.84 	 & 0.0290 	& 0.80 	 & 0.84 	 \\
\cline{1-16}
         \multicolumn{16}{|c|}{ $ds = 2000$} \\
        \cline{1-16}
DS1 & 0.0212 	& 51,226 	 & 558,352 	 & \textbf{0.0203} 	& 1.04 	 & 0.80 	 & 0.0227 	& 0.98 	 & 0.73 	 & 0.0249 	& 0.98 	 & 0.69 	 & 0.0245 	& 1.02 	 & 0.57 	 \\
DS2 & 0.1104 	& 43,823 	 & 782,890 	 & \textbf{0.1069}	& 0.98 	 & 0.65 	 & 0.1098 	& 0.90 	 & 0.54 	 & 0.1125 	& 0.92 	 & 0.50 	 & 0.1469 	& 0.80 	 & 0.39 	 \\
DS3 & 0.2351 	& 38,864 	 & 631,628 	 & 0.1505 	& 1.06 	 & 0.85 	 & \textbf{0.1198} 	& 1.01 	 & 0.79 	 & 0.1828 	& 0.94 	 & 0.73 	 & 0.2988 	& 0.94 	 & 0.62 	 \\
DS1D & \textbf{0.0107} 	& 59,385 	 & 598,296 	 & 0.0153 	& 0.91 	 & 0.77 	 & 0.0187 	& 0.89 	 & 0.67 	 & 0.0192 	& 0.95 	 & 0.65 	 & 0.0122 	& 0.85 	 & 0.54 	 \\
DS2D & 0.0911 	& 42,084 	 & 677,167 	 & \textbf{0.0868} 	& 0.97 	 & 0.79 	 & 0.0912 	& 0.99 	 & 0.66 	 & 0.0960 	& 0.98 	 & 0.63 	 & 0.1146 	& 0.89 	 & 0.46 	 \\
DS3D & \textbf{0.1215} 	& 41,138 	 & 653,667 	 & 0.1557 	& 0.95 	 & 0.80 	 & 0.2212 	& 0.89 	 & 0.73 	 & 0.1391 	& 0.91 	 & 0.73 	 & 0.2959 	& 0.81 	 & 0.61 	 \\
DS4 & \textbf{ 0.0155} 	& 571995 	 & 221648 	 & 0.0163 	& 0.88 	 & 0.95 	 & 0.0161 	& 0.89 	 & 0.95 	 & 0.0170 	& 0.76 	 & 0.90 	 & 0.0170 	& 0.76 	 & 0.90 	 \\
DS5 &  \textbf{0.0153} 	& 545483 	 & 223149 	 & 0.0158 	& 0.87 	 & 0.93 	 & 0.0164 	& 0.87 	 & 0.94 	 & 0.0171 	& 0.74 	 & 0.88 	 & 0.0171 	& 0.74 	 & 0.88 	 \\
TP1 & \textbf{0.0156} 	& 6,019 	 & 180,777 	 & 0.0163 	& 0.99 	 & 0.96 	 & 0.0167 	& 0.97 	 & 0.96 	 & 0.0183 	& 0.96 	 & 0.92 	 & 0.0183 	& 0.96 	 & 0.92 	 \\
TP2 & 0.0243 	& 8,277 	 & 179,399 	 & 0.0248 	& 0.97 	 & 0.95 	 & \textbf{0.0232}	& 0.96 	 & 0.93 	 & 0.0254 	& 0.87 	 & 0.91 	 & 0.0259 	& 0.87 	 & 0.89 	 \\
\cline{1-16}
         \multicolumn{16}{|c|}{ $ds = 5000$} \\
        \cline{1-16}
DS1 & 0.0235 	& 49,828 	 & 605,486 	 & 0.0179 	& 1.22 	 & 0.87 	 & \textbf{0.0168} 	& 1.11 	 & 0.77 	 & 0.0245 	& 1.10 	 & 0.74 	 & 0.0187 	& 1.08 	 & 0.65 	 \\
DS2 & 0.1044 	& 43,081 	 & 805,119 	 & \textbf{0.0994}	& 0.98 	 & 0.75 	 & 0.1266 	& 0.98 	 & 0.63 	 & 0.1114 	& 0.99 	 & 0.62 	 & 0.1130 	& 0.90 	 & 0.47 	 \\
DS3 & 0.2044 	& 42,240 	 & 681,821 	 & \textbf{0.1262} 	& 1.01 	 & 0.86 	 & 0.1907 	& 0.99 	 & 0.83 	 & 0.1390 	& 1.03 	 & 0.80 	 & 0.2092 	& 0.89 	 & 0.70 	 \\
DS1D & 0.0201 	& 52,762 	 & 620,806 	 & 0.0177 	& 0.98 	 & 0.79 	 & 0.0183 	& 0.93 	 & 0.75 	 & 0.0179 	& 0.99 	 & 0.73 	 & \textbf{0.0156} 	& 0.98 	 & 0.63 	 \\
DS2D & \textbf{0.0914}	& 41,710 	 & 816,763 	 & 0.0977 	& 0.98 	 & 0.72 	 & 0.1181 	& 0.91 	 & 0.61 	 & 0.0949 	& 1.10 	 & 0.59 	 & 0.1020 	& 0.99 	 & 0.48 	 \\
DS3D & \textbf{0.1345} 	& 46,867 	 & 714,076 	 & 0.1606 	& 0.82 	 & 0.79 	 & 0.1523 	& 0.93 	 & 0.78 	 & 0.1648 	& 0.89 	 & 0.75 	 & 0.1660 	& 0.77 	 & 0.66 	 \\
DS4 &  \textbf{0.0123} 	& 1053485 	 & 300147 	 & 0.0124 	& 0.93 	 & 0.96 	 & 0.0123 	& 1.00 	 & 1.00 	 & 0.0124 	& 0.93 	 & 0.96 	 & 0.0124 	& 0.93 	 & 0.96 	 \\
DS5 &  \textbf{0.0125} 	& 988365 	 & 297304 	 & 0.0128 	& 0.93 	 & 0.96 	 & 0.0125 	& 1.00 	 & 1.00 	 & 0.0128 	& 0.93 	 & 0.96 	 & 0.0128 	& 0.93 	 & 0.96 	 \\
TP1 & \textbf{0.0128} 	& 8,017 	 & 230,387 	 & 0.0133 	& 0.92 	 & 0.96 	 & 0.0134 	& 1.00 	 & 0.97 	 & 0.0133 	& 0.92 	 & 0.96 	 & 0.0137 	& 0.92 	 & 0.94 	 \\
TP2 & 0.0237 	& 8,380 	 & 212,285 	 & \textbf{0.0234}	& 1.00 	 & 0.97 	 & \textbf{0.0234} 	& 1.00 	 & 0.98 	 & \textbf{0.0234} 	& 1.00 	 & 0.97 	 & 0.0239 	& 1.00 	 & 0.95 	 \\
\hline
    \end{tabular}  
    \label{tab:keyparameter_raw}
\end{table*}

\clearpage

\section{Experiments on deceptive version of DS4D and DS5D}
In this section, we show the performance of PSP-BLEMO on deceptive version of DS4 and DS5 problem formulated in \cite{bing2024BL}, referred to as DS4D and DS5D. The deceptiveness is introduced into the problems by integrating $\{\mathbf{x}_{l(K+1)},\ldots \mathbf{x}_{l(K+L)}\}$ into the UL objective. We compare the performance with cG-BLEMO to show demonstrate the performance for cases where the problem has deceptiveness and VAA simultaneously.

\begin{figure*}[!ht]
    \centering
    \subfloat[DS4D]  
    {\includegraphics[width=0.35\textwidth]{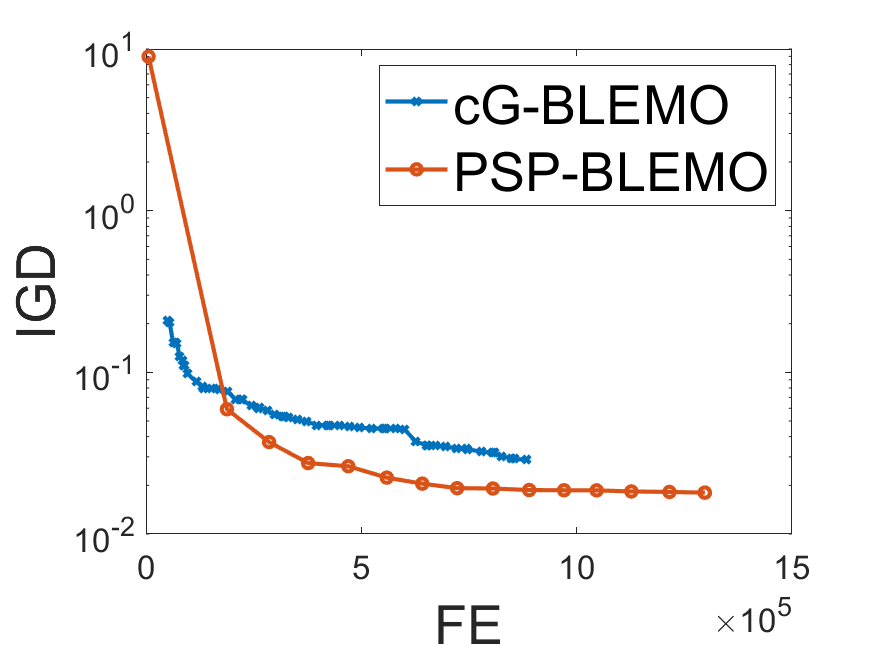}
         \label{fig:ds4d_converge_transfer} }           
    \subfloat[DS5D]
    {\includegraphics[width=0.35\textwidth]{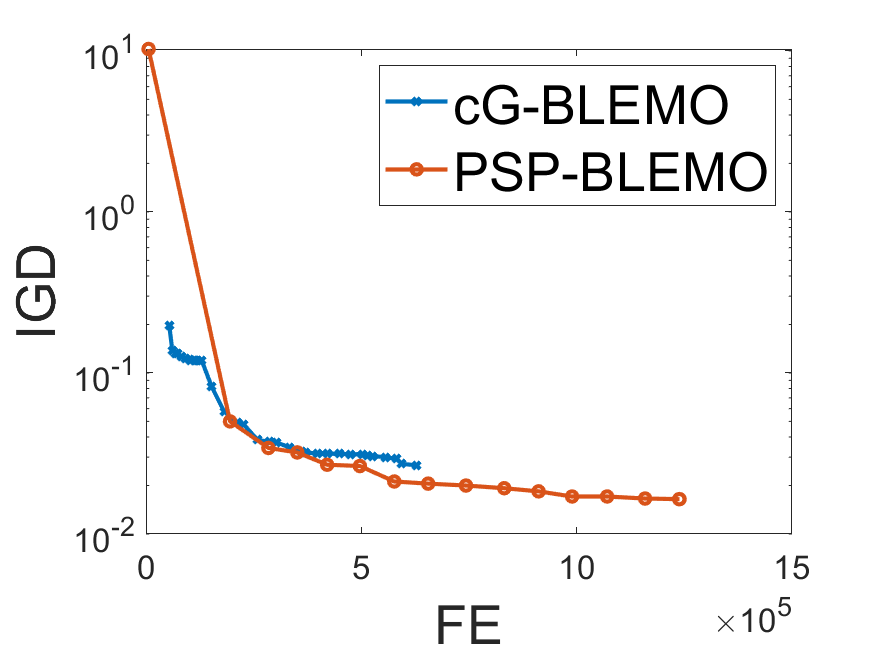}}
        \label{fig:ds5d_converge_transfer}
    \caption{Median convergence plots}
    \label{fig:DS45_converge}
\end{figure*}

In Fig.~\ref{fig:DS45_converge}, median convergence plots are shown between PSP-BLEMO and cG-BLEMO. It is evident that PSP-BLEMO can converge to much lower IGD values. There is a notable drop in IGD for PSP-BLEMO, which corresponds to the change in IGD in the first UL generation before and after additional UL search, as described by Algo.5 in main manuscript. Additionally, it can be observed that PSP-BLEMO requires significantly more function evaluations (FEs) to meet the stopping condition. To align FE consumption levels, we truncate PSP-BLEMO's run to match the FE consumption of cG-BLEMO.  
To do this, the runs for both algorithms are sorted based on the FE consumption,and then PSP-BLEMO metric (IGD value) is extracted at the point of corresponding cG-BLEMO FE consumption. 
The truncated IGD values are shown in the third column of Table~\ref{tab:correction_igd}. The first two columns of Table ~\ref{tab:correction_igd} show the original mean and standard deviation of search results. For the original results~(with no truncation), we can see that PSP-BLEMO performs significantly better across in two deceptive problems. Even when truncating the FE levels, PSP-BLEMO still performs better. The high IGD values of cG-BLEMO for DS4D and DS5D suggest that its UL search is misled by the deceptiveness of the problems. Overall, these experiments suggest that PSP-BLEMO can effectively handle problems with deceptiveness, demonstrating competitive performance.

\begin{table}[!ht]\footnotesize
    \centering
     \caption{IGD comparison between cG-BLEMO and PSP-BLEMO. Symbols $\uparrow_{(i)}$, $\downarrow_{(i)}$ and $\approx_{(i)}$ represents the statistical significance~(better, worse or equivalent) when compared to the algorithm $(i)$}
   \begin{tabular}{|l| r  r | r  r r | r  r  r|}
    \hline
         Prob. &  \multicolumn{2}{c|}{PSP-BLEMO$(1)$} & \multicolumn{3}{c}{cG-BLEMO$(2)$} & \multicolumn{3}{|c|}{PSP-BLEMO truncated} \\
         \cline{2-9}
               & mean & std & mean & std & sig & mean & std & sig\\
         \cline{1-9} 
         DS4D& 0.0187 	& 0.0027 	& 28.6090 	& 117.0015 &	$\downarrow_1$ &         0.0247 	& 0.0054 & $\uparrow_2$\\
        DS5D& 0.0175 	& 0.0023 	& 221.6651 	& 948.1017 &	$\downarrow_1$ &
        0.0297 	& 0.0129 & $\uparrow_2$ \\
         \hline
    \end{tabular}
    \label{tab:correction_igd}
\end{table}

\clearpage

\section{FE consumption with additional stopping condition}

Further to the above experiments, we evaluate the performance in the manner suggested in the recent study proposing cG-BLEMO~\cite{Conditional_wang2023}. 
Besides HV stopping condition, PSP-BLEMO, MOBEA-DPL,cG-BLEMO and SVM-BLMO are terminated when their UL IGD values are better than BLMOCC's reported IGD values. Thus with IGD values on the similar level, one can observe how swiftly the algorithms can achieve it. From Table~\ref{tab:compare_FE_igdstop}, we can see that PSP-BLEMO uses fewer FEs on UL and LL in most of the problems.

\begin{table*}[!ht]\footnotesize
\caption{FE consumption using BLMOCC reported IGD as the additional stopping condition. Algorithms stops when its IGD value decreases lower than corresponding reported mean value of BLMOCC} 
    \centering
    \tabcolsep 0.4mm
    \renewcommand\arraystretch{0.8}
    \begin{tabular}{|l| r r r  |r r r | r r r| r r r|r r r|r r r|}
    \hline
         \multirow{3}{*}{Prob.}& \multicolumn{18}{|c|}{UL FE} \\
        \cline{2-19}
         & \multicolumn{3}{|c}{PSP-BLEMO} & \multicolumn{3}{|c}{$\frac{\text{cG-BLEMO}}{\text{PSP-BLEMO}}
         $} & \multicolumn{3}{|c|}{$\frac{\text{MOBEA-DPL}}{\text{PSP-BLEMO}}
         $} & \multicolumn{3}{|c|}{$\frac{\text{BLMOCC}}{\text{PSP-BLEMO}}
         $} & \multicolumn{3}{c|}{$\frac{\text{H-BLEMO}}{\text{PSP-BLEMO}}
         $}&  \multicolumn{3}{c|}{$\frac{\text{SVM-BLEMO}}{\text{PSP-BLEMO}}
         $}\\
         \cline{2-19}
         &   Min & Med & Max & Min  & Med & Max & Min  & Med & Max &Min  & Med & Max&Min  & Med & Max&Min  & Med & Max\\
        \hline
DS1& 29,677 	& 37,927 	& 97,934 	& 2.43 	& 4.63 	& 4.26 	& 15.10 	& 19.12 	& 25.91 	& 1.20 	& 1.00 	& 0.44 	& 2.95 	& 2.42 	& 0.94 	& 0.06 	& 0.10 	& 0.06 	\\
DS2& 28,903 	& 39,550 	& 55,458 	& 1.77 	& 4.37 	& 5.85 	& 21.04 	& 29.13 	& 39.88 	& 1.61 	& 1.30 	& 1.04 	& 3.65 	& 2.95 	& 2.10 	& 0.08 	& 0.07 	& 0.08 	\\
DS3& 39,141 	& 48,443 	& 68,271 	& 0.30 	& 3.93 	& 5.37 	& 4.32 	& 6.63 	& 7.31 	& 1.27 	& 1.08 	& 0.80 	& 2.88 	& 2.45 	& 1.74 	& 0.02 	& 0.04 	& 0.02 	\\
DS4& 61,228 	& 73,319 	& 90,631 	& 0.39 	& 0.64 	& 1.44 	& 1.30 	& 1.86 	& 4.18 	& 0.36 	& 0.43 	& 0.41 	& 0.62 	& 0.73 	& 0.59 	& 0.07 	& 0.19 	& 0.07 	\\
DS5& 54,409 	& 68,154 	& 77,026 	& 0.47 	& 0.63 	& 1.20 	& 1.51 	& 1.87 	& 5.67 	& 0.54 	& 0.53 	& 0.59 	& 0.87 	& 0.83 	& 0.74 	& 0.07 	& 0.16 	& 0.07 	\\
TP1& 400 	& 780 	& 2,000 	& 6.95 	& 5.72 	& 4.39 	& 5.51 	& 4.46 	& 4.50 	& 21.40 	& 11.97 	& 5.13 	& 31.09 	& 18.16 	& 7.08 	& 1.00 	& 0.96 	& 1.00 	\\
TP2& 1,500 	& 1,960 	& 3,166 	& 9.89 	& 9.05 	& 7.73 	& 11.98 	& 17.88 	& 15.04 	& 7.24 	& 6.05 	& 4.18 	& 11.51 	& 9.56 	& 5.92 	& 0.86 	& 2.09 	& 0.86 	\\
        \hline
        \multirow{3}{*}{Prob.} & \multicolumn{18}{|c|}{LL FE} \\
        \cline{2-19}
         &  \multicolumn{3}{|c}{PSP-BLEMO} & 
            \multicolumn{3}{|c}{$\frac{\text{cG-BLEMO}}{\text{PSP-BLEMO}}$} & 
            \multicolumn{3}{|c|}{$\frac{\text{MOBEA-DPL}}{\text{PSP-BLEMO}}$}&
            \multicolumn{3}{|c|}{$\frac{\text{BLMOCC}}{\text{PSP-BLEMO}}$} & 
            \multicolumn{3}{c|}{$\frac{\text{H-BLEMO}}{\text{PSP-BLEMO}}$}&  
            \multicolumn{3}{c|}{$\frac{\text{SVM-BLEMO}}{\text{PSP-BLEMO}}$}\\
         \cline{2-19}
         &    Min & Med & Max & Min  & Med & Max & Min  & Med & Max &Min  & Med & Max &Min  & Med & Max&Min  & Med & Max\\
        \hline
      DS1& 535,225 	& 623,446 	& 795,528 	& 0.67 	& 1.39 	& 2.67 	& 5.99 	& 8.25 	& 18.09 	& 1.63 	& 1.64 	& 1.49 	& 5.27 	& 5.49 	& 4.30 	& 0.53 	& 0.85 	& 1.14 	\\
DS2& 570,626 	& 719,378 	& 1,131,016 	& 0.64 	& 1.81 	& 3.15 	& 10.34 	& 10.46 	& 11.15 	& 2.54 	& 2.32 	& 1.85 	& 7.86 	& 6.53 	& 4.15 	& 0.48 	& 0.50 	& 0.81 	\\
DS3& 980,487 	& 1,115,325 	& 1,431,745 	& 0.31 	& 2.91 	& 4.54 	& 4.68 	& 7.36 	& 8.55 	& 1.32 	& 1.38 	& 1.28 	& 4.05 	& 4.24 	& 3.30 	& 0.23 	& 0.51 	& 1.36 	\\
DS4& 47,489 	& 118,262 	& 118,382 	& 1.20 	& 0.87 	& 2.10 	& 6.67 	& 4.52 	& 12.80 	& 14.94 	& 6.36 	& 6.92 	& 28.57 	& 12.14 	& 12.12 	& 2.79 	& 1.43 	& 2.24 	\\
DS5& 118,105 	& 118,255 	& 118,362 	& 0.52 	& 0.78 	& 1.66 	& 2.73 	& 4.27 	& 14.35 	& 6.87 	& 7.63 	& 8.98 	& 14.11 	& 15.15 	& 15.14 	& 1.10 	& 1.34 	& 1.82 	\\
TP1& 118,137 	& 141,721 	& 199,441 	& 0.28 	& 0.33 	& 0.39 	& 0.26 	& 0.36 	& 0.65 	& 3.53 	& 3.16 	& 2.45 	& 4.70 	& 4.44 	& 3.16 	& 1.00 	& 0.84 	& 0.60 	\\
TP2& 142,868 	& 152,525 	& 178,823 	& 0.48 	& 0.57 	& 0.56 	& 0.91 	& 1.72 	& 1.96 	& 0.89 	& 1.00 	& 1.09 	& 1.85 	& 2.09 	& 1.79 	& 0.84 	& 0.87 	& 0.95 	\\
        \hline
        \end{tabular}   
        \label{tab:compare_FE_igdstop}
\end{table*}

\bibliographystyle{ACM-Reference-Format}
\bibliography{main}